\documentclass{article} 

\usepackage{amsmath,amsfonts,bm}

\newif\ifboldmatrix
\boldmatrixtrue

\ifboldmatrix\newcommand{\boldmatrix}[1]{\mathbf{#1}}\else\newcommand{\boldmatrix}[1]{#1}\fi
\ifboldmatrix\else\fi
\ifboldmatrix\else\fi
\newcommand{\ba}{\ensuremath{\mathbf{a}}}

\newcommand{\bc}{\ensuremath{\mathbf{c}}}

\newcommand{\bmm}{\ensuremath{\mathbf{m}}}

\newcommand{\bp}{\ensuremath{\mathbf{p}}}

\newcommand{\bq}{\ensuremath{\mathbf{q}}}

\newcommand{\bt}{\ensuremath{\mathbf{t}}}

\newcommand{\bw}{\ensuremath{\mathbf{w}}}

\newcommand{\by}{\ensuremath{\mathbf{y}}}
\newcommand{\bz}{\ensuremath{\mathbf{z}}}

\newcommand{\bT}{\ensuremath{\boldmatrix{T}}}

\def\eqref#1{equation~\ref{#1}}

\def\1{\bm{1}}

\DeclareMathAlphabet{\mathsfit}{\encodingdefault}{\sfdefault}{m}{sl}
\SetMathAlphabet{\mathsfit}{bold}{\encodingdefault}{\sfdefault}{bx}{n}

\usepackage{arxiv}

\usepackage{hyperref}
\usepackage{url}
\usepackage{xspace}
\usepackage{microtype}
\usepackage{graphicx}
\usepackage{subfigure}
\usepackage{booktabs} 
\usepackage{multirow}
\usepackage{xspace}
\usepackage{amsmath}
\usepackage{amssymb}
\usepackage{bm}
\usepackage{listings}
\usepackage{caption}
\usepackage{tcolorbox}
\usepackage{cleveref}
\usepackage{hyperref}
\usepackage{cleveref}
\usepackage{longtable}
\usepackage{array}
\usepackage{xcolor}
\usepackage{xcolor} 
\usepackage{authblk}
\usepackage{natbib}
\usepackage{wrapfig}

\clearpage

\newtcolorbox{exampleboxcode}{colback=black!5, colframe=black!50, arc=4pt, boxrule=1pt,   left=5pt, right=5pt, top=0pt, bottom=0pt 
}

\definecolor{codegreen}{rgb}{0,0.6,0}
\definecolor{codegray}{rgb}{0.5,0.5,0.5}
\definecolor{codepurple}{rgb}{0.58,0,0.82}
\definecolor{backcolour}{rgb}{0.97,0.97,0.97}

\lstdefinestyle{codestyle}{
    commentstyle=\color{codegreen},
    keywordstyle=\color{magenta},
    numberstyle=\tiny\color{codegray},
    stringstyle=\color{codepurple},
    basicstyle=\ttfamily\scriptsize,
    breakatwhitespace=false,         
    breaklines=true,                 
    captionpos=b,                    
    keepspaces=true,                 
    numbers=left,                    
    numbersep=5pt,                  
    showspaces=false,                
    showstringspaces=false,
    showtabs=false,                  
    tabsize=2,
    xleftmargin=10pt,                
    xrightmargin=10pt                
}
\title{Minions: Cost-efficient Collaboration Between On-device and Cloud Language Models}

\newcommand{\system}{\textsc{Minion}$\mathcal{S}$\xspace}
\newcommand{\naive}{\textsc{Minion}\xspace}
\newcommand{\llama}{\textsc{Llama}\xspace}
\newcommand{\llamathree}{\textsc{Llama-3B}\xspace}
\newcommand{\llamaeight}{\textsc{Llama-8B}\xspace}
\newcommand{\llamaone}{\textsc{Llama-1B}\xspace}
\newcommand{\llamathreetwo}{\textsc{Llama-3.2}\xspace}
\newcommand{\llamathreeone}{\textsc{Llama-3.1}\xspace}

\newcommand{\qwenthree}{\textsc{Qwen-3b}\xspace}
\newcommand{\qwenseven}{\textsc{Qwen-7b}\xspace}
\newcommand{\gpt}{\textsc{GPT-4o}\xspace}
\newcommand{\qwen}{\textsc{Qwen2.5}\xspace}
\newcommand{\qasper}{\textsc{QASPER}\xspace}
\newcommand{\longhealth}{\textsc{LongHealth}\xspace}
\newcommand{\finance}{\textsc{FinanceBench}\xspace}
\newcommand{\locallm}{\mathrm{LocalLM}\xspace}
\newcommand{\remotelm}{\mathrm{RemoteLM}\xspace}
\newcommand{\systemmath}{\textsc{MinionS}\xspace}

\begin{document}

\renewcommand\Authands{, } 

\newcommand{\equalcontrib}{\textsuperscript{*}}
\newcommand{\affilmark}[1]{\textsuperscript{#1}} 

\author[1]{Avanika Narayan\equalcontrib}
\author[1,2,3]{Dan Biderman\equalcontrib}
\author[1]{Sabri Eyuboglu\equalcontrib}
\author[5]{Avner May}
\author[2,3]{\\Scott Linderman}
\author[4]{James Zou}
\author[1]{Christopher Ré}

\affil[1]{Department of Computer Science, Stanford University}
\affil[2]{Department of Statistics, Stanford University}
\affil[3]{Wu Tsai Neurosciences Institute, Stanford University}
\affil[4]{Departemnet of Biomedical Data Science, Stanford University}
\affil[5]{Together AI}

\affil[ ]{\texttt{\{avanikan,biderman,eyuboglu\}@stanford.edu}}

\date{}

\maketitle

\let\thefootnote\relax\footnotetext{\textsuperscript{*} Corresponding authors; equal contribution and random ordering for AN, SE, DB.}







\begin{abstract}
We investigate an emerging setup in which a small, on-device language model (LM) with access to local data communicates with a frontier, cloud-hosted LM to solve real-world tasks involving financial, medical, and scientific reasoning over long documents. 
\textit{Can a local-remote collaboration reduce cloud inference costs while preserving quality?} 
First, we consider a naïve collaboration protocol where the local and remote models simply chat back and forth. 
Because only the local model reads the full context, this protocol achieves a $30.4\times$ reduction in remote costs, but recovers only 87\% of the performance of the frontier model.
We identify two key limitations of this protocol:  the local model struggles to (1) follow the remote model's multi-step instructions and (2) reason over long contexts. 
Motivated by these observations, we study an extension of this protocol, coined \system, in which the remote model decomposes the task into easier subtasks over shorter chunks of the document, that are executed locally in parallel. \system reduces costs by $5.7\times$ on average while recovering 97.9\% of the performance of the remote model alone.
Our analysis reveals several key design choices that influence the trade-off between cost and performance in local-remote systems.
\end{abstract}


\section{Introduction}

\label{sec:intro}
Today's cloud-hosted frontier Language Models (LMs) can perform \textit{data-intensive reasoning}: they can generate and refactor code across entire repositories and make decisions based on financial, legal, and medical documents. However, accessing these models is expensive: processing a standard million-token code repository with OpenAI's o1 API costs $>\$15$ per query. 
At the same time, smaller LMs (1-8B parameters) are rapidly improving and can now run on personal computers (\href{https://ollama.com/}{Ollama}, \href{https://github.com/ggerganov/llama.cpp}{llama.cpp}) and smartphones \citep{mehta2024openelm,yi2024phonelm,xu2024empowering}. 
%
Yet, today, these small, on-device LMs are used mostly for \href{https://machinelearning.apple.com/research/introducing-apple-foundation-models}{simple tasks} such as tone adjustment and text completion~\citep{gunter2024apple}. They do not play a role in data-intensive reasoning tasks.

Inspired by the growing literature on multi-agent systems~\citep{wang2024mixture,guo2024large}, in this work we ask: \textit{how can a small, on-device LM collaborate with a frontier LM in the cloud to reduce inference costs on data-intensive reasoning tasks?} In particular, we study the \textit{communication protocols} that govern how the two LMs talk to each other, focusing on the tradeoff between cost and accuracy.
To mimic realistic use cases, we study tasks that involve varying levels of reasoning over large volumes of medical, financial, and academic data ~\citep{islam2023financebench, adams2024longhealth, dasigi2021dataset}.
%




\begin{figure*}[t]  
    \centering
    \includegraphics[width=\textwidth]{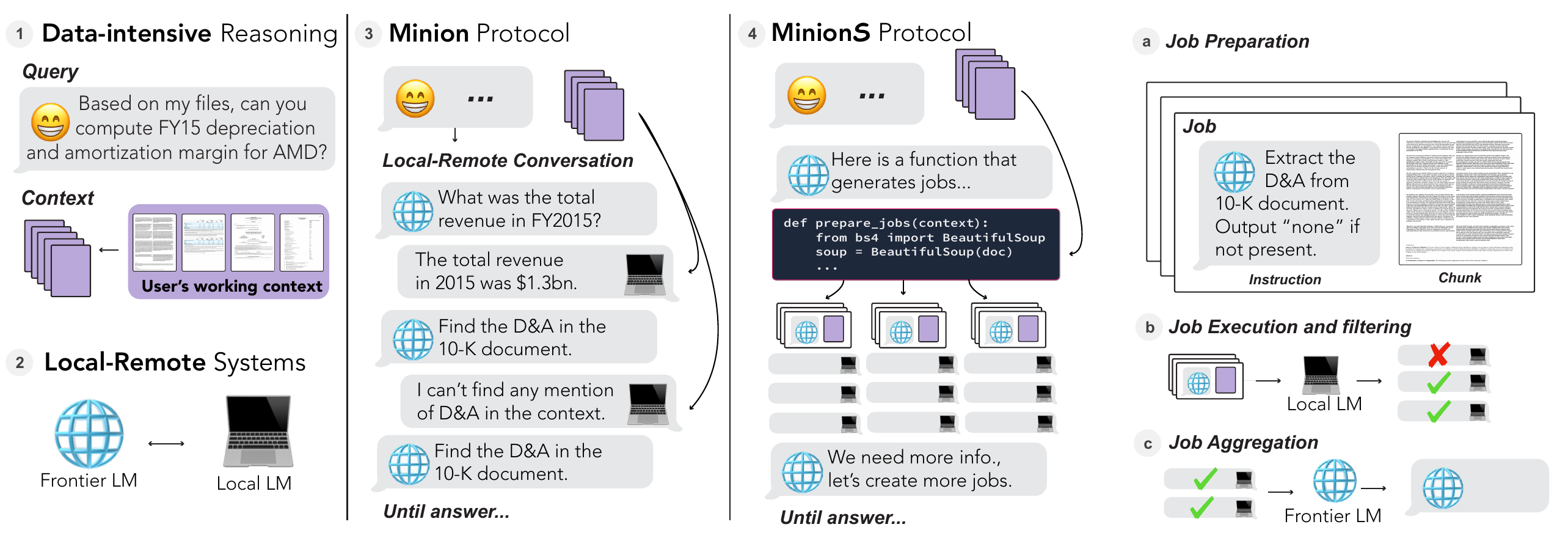}

    \caption{
        \textbf{Local-Remote Systems}. \naive and \system protocols. \textbf{(Left)} Problem set-up: local and remote LM collaborate on a data-intensive reasoning task. \textbf{(Center)} \naive: A simple communication protocol in which the local and remote models have an ``unconstrained'' back and forth chat. \textbf{(Right)} \system: an extension of \naive where the remote LM decomposes a query into many \textit{jobs} that are processed in parallel by the local model. Each job is a single-step instruction over a chunk of the context.
    }
    \label{fig:banner}

\end{figure*}

As our first attempt, we study a simple communication protocol we call \naive: an unconstrained chat between the local and remote models. \naive reduces cloud costs by only ``reading'' the data locally, and communicating a compressed version of the context to the remote model. 
We show that while \naive achieves a $30.4\times$ reduction in remote model costs, it trails behind the remote-only baseline by 9.4 accuracy points on average (with an 8B local model; see \Cref{sec:naive} for details).
%
In isolated ablations, we identify two key limitations of small LMs that hinder \naive's performance (Section~\ref{sec:naive}):
\begin{itemize}
        \item \emph{Small LMs struggle to follow multi-step instructions.} We find that splitting complex instructions into separate requests improves performance by $56\%$.
        \item \emph{Small LMs get confused by long contexts}. Increasing context length from $<1K$ to $>65K$ decreases performance by $13\%$ on a simple extraction task. 
\end{itemize}
Motivated by these limitations, we propose \system, an extension of \naive where the remote LM decomposes the problem into a set of \emph{single-step instructions} to be performed on smaller \emph{chunks} of the document.
Crucially, the remote model has to do this \textit{without} reading the full document, which it achieves by \textit{generating code} that is later executed locally where the document is.
%
%
%
%
More precisely, \system involves a loop over three steps:
\begin{enumerate}
    \item \textbf{Decompose}: Given a task, the remote model writes code that decomposes it into ``bite-sized'' \textit{subtasks}.
    \item \textbf{Execute}: The local LM then executes the subtasks in parallel and sends a filtered selection of the responses back to the remote model.
    \item \textbf{Aggregate}: The remote model aggregates the local outputs and either finalizes the solution or loops back to the Decompose step.
\end{enumerate}



\begin{wrapfigure}{r}{0.5\linewidth}

   \centering
    \includegraphics[width=\linewidth]{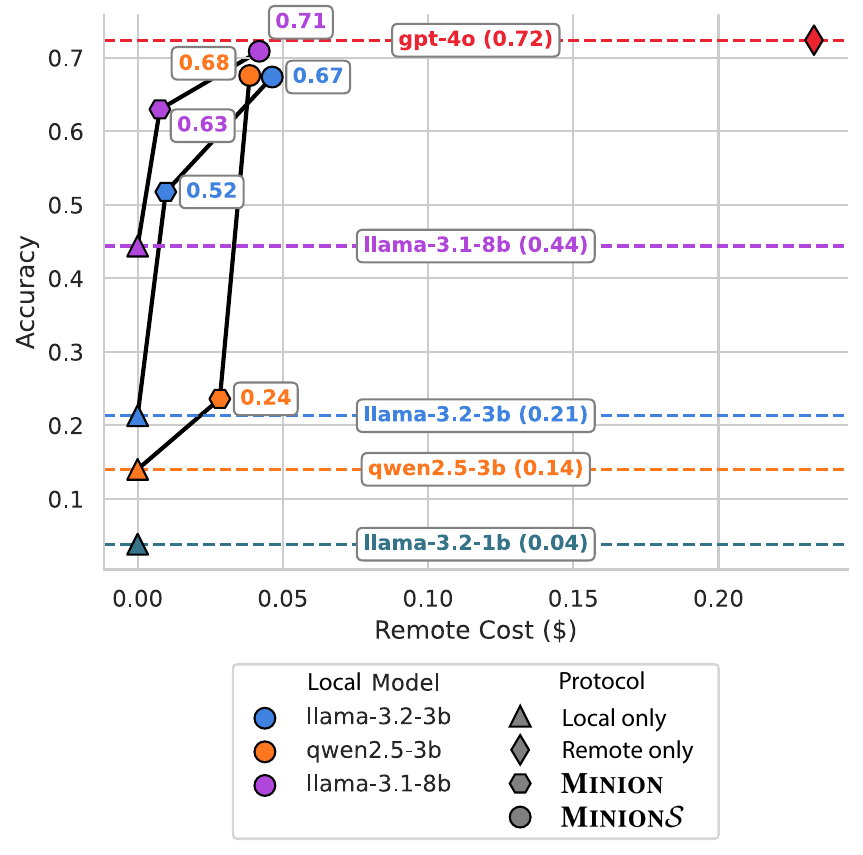}

    \caption{\textbf{Cost-Accuracy Tradeoff in Edge-Remote Systems.} 
    Macro-average accuracy ($y$-axis) vs. cost ($x$-axis) across \finance~\citep{islam2023financebench}, \longhealth~\citep{adams2024longhealth}, and \qasper~\citep{dasigi2021dataset}.     
    Accuracy represents the fraction of correct predictions, while cost is the average USD per query based on \gpt rates (Jan 2025: \$2.50/1M input tokens, \$10.00/1M output tokens); see \cref{sec:prelim}. 
    The table compares \naive (\cref{sec:naive}) and \system (\cref{sec:methods}) protocols against local-only and remote-only baselines. 
    Points, colored by local model, use \gpt as the remote model. Exact metrics in Table~\ref{table:main-tradeoff}.}
    \label{fig:main-tradeoff}

\end{wrapfigure}

Averaged across tasks, \system with an 8B parameter local LM can recover $97.9\%$ of the performance of remote-only systems at $18.0\%$ of the cloud cost (see Figure~\ref{fig:main-tradeoff}). 
With a 3B parameter local LM, \system  achieves $93.4\%$ of the performance of remote-only systems at $16.6\%$ of the cloud cost (see Figure~\ref{fig:main-tradeoff}).





We perform a detailed analysis of the design and hyperparameter space of \system. 
Our analysis highlights several ``knobs'' that allow us to trade off cost for quality. 

\textbf{(a) Model choice} \textit{How does the size and family of the language models affect the cost and quality?} We show that \system would not have been feasible until mid-2024 (due to the  the release of \textsc{gpt4-turbo} and \llamathreeone) and is now performant with the latest 3B-parameter models running locally.


\textbf{(b) Scaling parallel workloads on-device.} \textit{How should we structure parallel workloads at the edge to maximize performance?}  In \Cref{subsec:results-workloads}, we study three different strategies for increasing the parallel workload on-device: (a)  repeated sampling, (b) decomposition, and (c) context chunking. 
  We show that all three can independently improve quality at the expense of increased remote cost. 

\textbf{(c) Sequential communication protocols.} \textit{Can multiple rounds of communication improve quality? At what cost?} 
  We show that by increasing the number of sequential rounds of communication, we can  pay more to improve quality. 

To summarize, our main contributions are as follows:
\begin{itemize}
    \vspace{-0.5em}\item Propose \naive, a na\"{\i}ve local-remote LM communication protocol, that achieves $30.4\times$ efficiency over remote-only workloads while recovering $87\%$ of performance.
    \vspace{-0.5em}\item Propose \system, an extension that overcomes the limitations we identify in \naive, achieving $5.7\times$  cost-reduction over remote-only workloads and recovering $97.9\%$ of performance.
    \vspace{-0.5em}\item Conduct an in-depth analysis of \system, exploring design choices to traverse the cost-accuracy trade-off. 
    
\end{itemize}

\vspace{-1em}
\section{Related Work}

\label{sec:related-work}

\textit{See \Cref{app:related-work} for an extended discussion of related work.}

This study is inspired by a large body of work that explores how to combine multiple LMs and tools to improve quality and reduce cost of cloud workloads.
These include:

\setlength{\leftmargini}{12pt} 
\begin{itemize}

\item \textbf{Retrieval-Augmented Generation (RAG)} RAG workflows append retrieved chunks to a large LM's prompt~\citep{lewis2020retrieval,karpukhin2020dense, lee2019latent}, mitigating hallucinations and externalizing knowledge. We differ by assigning arbitrary subtasks to a local LM, which can \emph{execute} longer or more complex instructions on chunks before sending only the final extracted content. This helps reduce communication costs more than purely retrieving and passing raw text.

\item \textbf{Multi-LLM collaboration and routing} A growing body of work explores multi-agent or multi-LLM systems~\citep{guo2024large, wang2024mixture} and model-routing~\citep{chen2023frugalgpt, chen2024more}. Typically, these either combine multiple large models or choose one LM from a "menu" of models for the entire task. In contrast, we explicitly study a \emph{two-model collaboration} where the smaller local LM handles extensive on-device context, while the larger remote LM is called on selectively, reducing cloud inference costs.

\item \textbf{Compound LM systems} A broader line of research integrates LMs with retrieval modules, tool use, or orchestrators~\citep{saad2024archon, khattab2023dspy}. While they optimize accuracy or adapt prompts, they do not usually focus on asymmetric edge-cloud costs or local parallelization.

\end{itemize}

The specific techniques used in \system build upon several important ideas proposed in the literature:

\setlength{\leftmargini}{12pt} 
\begin{itemize}

\item \textbf{Orchestration for long-contexts} Prior works have used ``divide-and-conquer'' strategies to process long documents in smaller chunks~\citep{zhang2024chain, zhou2024llm, shankar2024docetl}. They define protocols for chunking, processing, and recombining content, sometimes with automated pipeline optimizations. However, these methods typically rely on a single large LM or multiple equally capable LMs, rather than an asymmetric local-remote collaboration with explicit cost constraints. They also do not explore parallel on-device tasks or multi-round communication with a cloud model. Other approaches improve single-LM handling of lengthy inputs by compressing, summarizing, or streaming data. Techniques like MemGPT~\citep{packer2023memgpt}, PRISM~\citep{jayalath2024long}, and writing-in-the-margins~\citep{russak2024writing} store partial results or structured data across external memory.  While such approaches reduce context overhead for a single LM, they do not address distributing computation across a local-remote system with distinct cost models.

\item \textbf{Decomposition techniques} The decomposition techniques used in \system are inspired by prior work showing how prompting for decomposition can improve small LM quality~\citep{arora2022ask, patel2022question, wu2022ai}. The \system protocol also builds upon the idea of using code to facilitate reasoning~\citep{arora2023evaporate,li2023chain}.

\item \textbf{Test-time sampling and verification} In \system we pair repeated test-time sampling on-device with verification in the cloud. This technique is motivated by extensive literature demonstrating the promise of using test-time sampling and verification to improve reasoning capabilities~\citep{brown2024large,song2024good,hassid2024larger,snell2024scaling,wu2024empirical}.

\end{itemize}


Some recent works have explored aspects of the local-remote setting.
Several study how local-remote systems can limit leakage of private information to a cloud-hosted LM API~\citep{siyan2024papillon, zhang2024cogenesis}. 
In this work, we do not address privacy concerns, though these privacy techniques 
can be used in conjunction with \system.
Other techniques partition LM layers between local and cloud devices~\citep{jin2024collm, yang2024perllm} without a multi-round dialogue. Our system is distinct in that the local LM and remote LM \emph{collaborate in natural language} on tasks that draw on a large private context. This two-model interplay underlies our focus: reducing cloud inference costs while preserving performance.

\vspace{-0.75em}
\section{Preliminaries}
\label{sec:prelim}
\label{sec:prelim-setup}
We study the tradeoff between the \textit{quality} of a local-remote system and the \textit{cost} of running it. We first outline the problem setup and then provide details on how we measure accuracy and cost.

\vspace{-0.5em}\paragraph{Problem setup} We study language tasks that involve a context $\bc$ (\textit{e.g.} a long document), a query $\bq$ against that context, and a ground-truth answer $\by$ (see (1) in ~\Cref{fig:banner}).
\newtcolorbox{examplebox}{colback=blue!5!white, colframe=blue!75!black}

\begin{examplebox}
\small
\textbf{Context ($\bc$):} \texttt{The Fiscal Year 2015 10-K report for Advanced Micro Devices, Inc.}

\textbf{Query ($\bq$):} \textit{
    Compute the 2015 depreciation and amortization margin for AMD (in percentage). 
}

\textbf{Answer ($\by$):} \texttt{US\$394,328 million}
\end{examplebox}

A \textit{local-remote} system $\mathrm{S}$ (see (2) in Figure~\ref{fig:banner}), consists of two language models that must collaborate to solve the task---a small LM ($\locallm$) running on on-device, and a large frontier LM ($\remotelm$) running in the cloud. 
$\mathrm{S}$ ingests a context and a query, and applies both models in conjunction to output a predicted answer: $\hat{\by} \sim \mathrm{S}(\bc, \bq)$. 

\vspace{-0.5em}\paragraph{Measuring quality} We evaluate the performance of $\mathrm{S}$ on a dataset $\mathcal{D} = \{(\bc_i, \bq_i, \by_i)\}_{i=1}^N$, via a scoring metric $s(\hat{\by_i}, \by_i)$. Here, $s(\cdot, \cdot)$ is binary (correct/incorrect) and we report \textit{accuracy}. As baselines, we compare $\mathrm{S}$ to $ \hat{\by}_{\text{remote}}\sim \remotelm(\bc, \bq)$ and $\hat{\by}_{\text{local}} \sim \locallm(\bc, \bq)$.

\vspace{-0.5em}\paragraph{Measuring cost}

Monetary cost (in \$USD) is our primary cost metric. We assume that $\remotelm$ calls incur a cost while $\locallm$ calls are free, ignoring the fixed cost of the hardware and marginal cost of energy consumption. 

More concretely, the cost of calls to $\remotelm$ are proportional to a weighted sum of the number of prefill (\textit{i.e.} input) tokens and decode (\textit{i.e.} output) tokens:
\begin{equation*}
    C_{\text{remote}}(n_{\text{prefill}}, n_{\text{decode}}) \propto n_{\text{prefill}} + \alpha \cdot n_{\text{decode}},
\end{equation*}
where $\alpha$ varies by provider ($\approx$1-- 5)~\citep{dubey2024llama3,anthropic2024claude}.
Decode tokens are more expensive since the decoding stage is IO-bound~\citep{leviathan2023fast} and has lower GPU utilization. This is because generating each decode token requires loading the full model and KV cache into GPU SRAM.
    

In this work we do not focus on optimizing the latency of local-remote systems.
However, in \cref{app:edgecost} we show analytically that there are important regimes where the systems proposed (\naive, \system) incur at most a $5\times$ increase in latency relative to performing the entire operation remotely.
This is possible because these systems avoid the costly step of processing the entire document with the huge $\remotelm$, and because they can make efficient use of the local hardware by batching (\textit{e.g.} in \system).
We leave a detailed empirical study of the latency trade-offs of these local-remote systems for future work.

\vspace{-0.75em}
\section{\naive: A naïve communication protocol}
\label{sec:naive}
\vspace{-0.5em}
\begin{wrapfigure}{r}{0.5\linewidth}
    \centering
    \includegraphics[width=\linewidth]{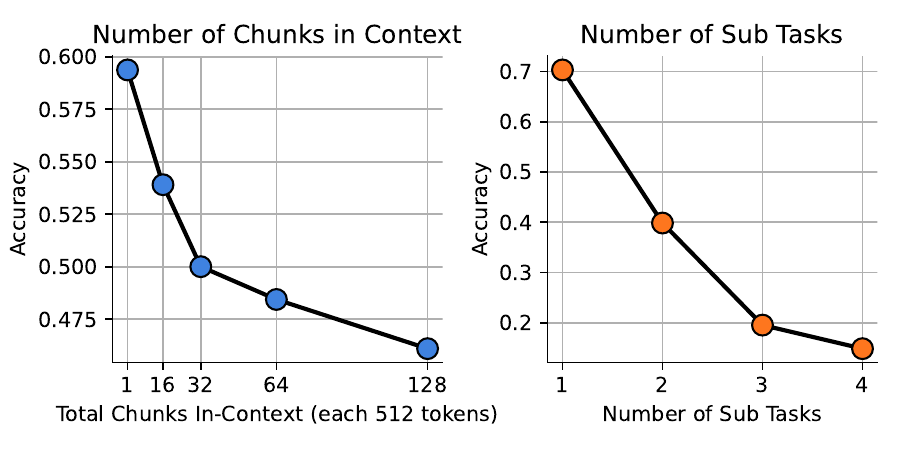}

    \caption{\textbf{Analysis of small LM limitations}. Evaluation of \llamathreetwo-3B on simple extraction tasks (see Section~\ref{app:naive-analysis}). \textbf{(Left)} Performance drops significantly as context length increases. \textbf{(Right)} Increasing sub-task complexity reduces performance, with fewer sub-tasks yielding better results.}
    \label{fig:small-micros}
\end{wrapfigure}

In this section, we describe \naive, a baseline local-remote communication protocol, which implements a simple free-form conversation between $\locallm$ and $\remotelm$.

It begins with system prompts for both models informing them of the query $\bq$ and that they will be collaborating with another model to answer it (see (3) in Figure~\ref{fig:banner}). 
Crucially, the system prompt for the $\locallm$ includes the full context $\bc$ while the system prompt for the $\remotelm$ does not.
After the system prompts, the two models chat back and forth with one another until the $\remotelm$ provides a final answer to the query. 
%
%
\textit{See \Cref{app:methods-naive} for a detailed description of the \naive protocol.}

We compare \naive to a baseline where $\remotelm$ is given the full context and the query. Excitingly, \naive reduces $\remotelm$ costs by $38.13\times$, $31.3\times$, and $20.9\times$ on \finance, \longhealth and \qasper, respectively (see ~\Cref{subsec:exp-setup} for dataset details).
Averaged across these datasets, it closes $87.0\%$ of the quality gap between $\remotelm$ and $\locallm$ operating alone. 

To further close the gap, we analyze \naive conversations and find that in unconstrained chat, $\remotelm$ often gives $\locallm$ \emph{complicated instructions} over \emph{long contexts}. \Cref{app:naive-analysis} presents micro-experiments illustrating $\locallm$'s struggles with these instructions:
\begin{enumerate}
    \item \textbf{$\locallm$ struggles to handle multi-step instructions.} Using \gpt, we generate instructions with varying numbers of sub-parts. We then show splitting sub-parts into separate requests leads to a 56 point performance improvement (see Figure~\ref{fig:small-micros}).
    
    \item \textbf{$\locallm$ struggles to reason across long contexts}. We show how increasing context length from $<1K$ to $>65K$ tokens can decrease performance by 13\% on a simple extraction instruction (see Figure~\ref{fig:small-micros}).
\end{enumerate}

Put simply, these smaller LMs are currently better equipped to answer simple queries on shorter contexts.

\vspace{-1em}\section{\system: A decomposition-based communication protocol}
\label{sec:methods}


Motivated by these observations, we introduce \system, a simple extension of the naïve communication protocol discussed in \cref{sec:naive}.
\system uses a divide-and-conquer strategy where the $\remotelm$ decomposes the task into \textit{simpler} jobs that can be run \textit{in parallel} (see (4) in Figure~\ref{fig:banner}). 
Throughout this section we will continue with the example task introduced in \Cref{sec:prelim-setup}.


\vspace{-0.5em}
\subsection{Protocol description}
\system protocol is a loop over three steps:
\begin{enumerate}
    \vspace{-0.5em}\item \textit{Job preparation on remote.}  $\remotelm$ writes code that generates a list of job specifications for $\locallm$ (see 4(a) in Figure~\ref{fig:banner}).
    \vspace{-0.5em}\item \textit{Job execution and filtering locally.} The job specifications are executed locally with the $\locallm$, and outputs are filtered (see 4(b) in Figure~\ref{fig:banner}).
    \vspace{-0.5em}\item \textit{Job aggregation on remote.} The remote model receives the filtered outputs and decides whether to output an answer or begin another iteration (see 4(c) in Figure~\ref{fig:banner}).
\end{enumerate}

\paragraph{Step 1: Job preparation on remote.} 
In this step, the $\remotelm$ generates a list of jobs that the $\locallm$ will run in parallel. A \textit{job} is a specification of a subtask, which can be converted into a prompt and sent to the local model. More precisely, a job, $\bt$, is a context-instruction pair $\bt^{(i)} = (\tilde{\bq}^{(i)}, \tilde{\bc}^{(i)})$. We denote a list of jobs with $\bT = [ \bt^{(1)}, \bt^{(2)}, ... ]$

\begin{examplebox}
    \small \textbf{Instruction ($\tilde{\bq}^{(i)}$):} Extract the total revenue for FY2015, abstain if not present. Try to look for the income statement and make sure it is from 2015. \\
    \textbf{Context ($\tilde{\bc}^{(i)}$}): ``Operating income for North America for the years ended..." 
\end{examplebox}

Crucially, the context $\tilde{\bc}^{(i)}$ for a job need not include the entire context $\bc$ of the full task. 
In principle, this allows us to chunk the context into more manageable pieces, which can be executed in parallel. 
\textit{But how can the $\remotelm$ chunk the context without reading it?}

To avoid reading the entire context, we have the remote model generate a Python function, $\textbf{f}(\bc, \bT)$, that accepts the full task context $\bc$ and jobs from the last iteration $\hat{\bT}$ and outputs a new list of jobs $\bT$. Specifically, we prompt $\remotelm$ with the task query $\bq$ and instruction prompt $\bp_{\text{decompose}}$:
    $\textbf{f}(\cdot, \cdot) \sim \remotelm(\bq, \bp_{\text{decompose}})$. 
Then, on-device, the function is executed with the context $\bc$ as the argument producing a list of jobs $\bT = \textbf{f}(\bc, \hat{\bT})$.

This strategy, which builds on work using LMs to generate code for information extraction~\citep{arora2023evaporate,li2023chain}, allows us to decouple the number of unique jobs from the number tokens generated by the cloud model.
For example, the code below, which is an abbreviated version of a function that was generated by the cloud model, is less than fifteen lines but can generate hundreds of jobs. 
\begin{exampleboxcode}
\begin{lstlisting}[language=Python]
@dataclass
class Job:
    instruction: str
    chunk: str
    
def f(ctx: str, last_jobs: List[Job]) -> List[Job]:
    jobs = []
    instructions = ["Extract the total revenue for...", "In the statement of cash flow..."]
    chunks = chunk_on_pages(ctx) # chunk context into pages
    for chunk in chunks:
        for instr in instructions:
            for _ in range(5):
                jobs.append(Job(instr, chunk))
    return jobs
\end{lstlisting}
\end{exampleboxcode}

Additionally, by passing the previous iteration's jobs and responses $\bT$ (\texttt{last\_jobs} in the code snippet), the large model can create jobs which build on previous responses. For example, the cloud model in the second round might zoom in on a relevant chunk identified in the first round.
For more examples of generated functions or details on the exact prompt used to generate the code, see Appendix~\ref{app:methods-prompts}. 
    
\vspace{-0.5em}\paragraph{Step 2: Job execution and filtering on-device.} 
In this step, we convert the jobs $\bT = [ \bt^{(1)}, \bt^{(2)}, ... ]$ into prompts and execute them in parallel locally. 

The jobs are fed in batch(es) to the $\locallm$ together with a system prompt $\bp_{\text{worker}}$ that instructs the model to either abstain or return a JSON object $\bz^{(i)}$ with fields \texttt{explanation}, \texttt{citation}, and \texttt{answer} to help us verify its reasoning. 
\begin{equation}
    \bz^{(i)} \sim \locallm(\bt^{(i)},\bp_{\text{worker}})
\end{equation}

After the $\locallm$ has generated the results, we discard any $\bz^{(i)}$ for which the model abstained. Intuitively, many instructions will be irrelevant to their paired chunks, allowing the $\locallm$ to abstain and avoid sending unnecessary information to the $\remotelm$.
The surviving subtask-chunk pairs are aggregated to form the formatted string $\bw$. 

\textbf{Step 3: Job aggregation on remote.}  $\remotelm$ receives $\bw$ and a synthesis prompt $\bp_{\text{synthesize}}$, instructing it to generate a JSON object $\ba$ with a ``decision'' field for sufficiency and a ``response'' field for a (potential) final answer:
\begin{equation}
    \hat{\by} \sim \remotelm(\bw, \bp_{\text{synthesize}})
\end{equation}
If the $\remotelm$ decides that more information is needed, the loop continues from Step 1. 

There are several ways to maintain context across \system rounds. 
One simple approach is to keep the entire the conversation in context. 
However, this strategy incurs significant additional cost, even with prompt caching. 
We experiment with two alternatives: (1) \textit{simple retries}, in which only the $\remotelm$'s advice is carried over between rounds and (2) \textit{scratchpads}, in which the $\remotelm$ can record what it learned from the round before proceeding to the next. 

\begin{table*}[]
\centering
\scriptsize

\begin{tabular}{l l l c c c c c c c c}
\toprule
\multicolumn{1}{c}{\textbf{Protocol}} & \multicolumn{1}{c}{\textbf{Local Model}} & \multicolumn{1}{c}{\textbf{Remote Model}} & \multicolumn{2}{c}{Macro Avg.} & \multicolumn{2}{c}{\finance} & \multicolumn{2}{c}{\longhealth} & \multicolumn{2}{c}{\qasper} \\
 &  &  & Acc. & Cost & Acc. & Cost & Acc. & Cost & Acc. & Cost \\
\hline
Remote Only & --- & \gpt & \textbf{0.724} & \$0.233 & \textbf{0.826} & \$0.261 & \textbf{0.748} & \$0.301 & \underline{0.598} & \$0.137\\
\midrule
Local Only & \llamaeight & --- & 0.444 & \$0.000 & 0.326 & \$0.000 & 0.468 & \$0.000 & 0.538 & \$0.000\\
Local Only & \llamaone & --- & 0.038 & \$0.000 & 0.000 & \$0.000 & 0.115 & \$0.000 & 0.000 & \$0.000\\
Local Only & \llamathree & --- & 0.213 & \$0.000 & 0.130 & \$0.000 & 0.345 & \$0.000 & 0.164 & \$0.000\\
Local Only & \qwenthree & --- & 0.140 & \$0.000 & 0.087 & \$0.000 & 0.177 & \$0.000 & 0.156 & \$0.000\\
\midrule
\naive & \llamaeight & \gpt & 0.630 & \$0.008 & \underline{0.804} & \$0.007 & 0.635 & \$0.010 & 0.450 & \$0.007\\
\naive & \llamathree & \gpt & 0.518 & \$0.010 & 0.698 & \$0.010 & 0.482 & \$0.009 & 0.372 & \$0.011\\
\naive & \qwenthree & \gpt & 0.236 & \$0.028 & 0.217 & \$0.029 & 0.281 & \$0.021 & 0.210 & \$0.035\\
\midrule
\system & \llamaeight & \gpt & \underline{0.709} & \$0.042 & \underline{0.804} & \$0.053 & \underline{0.740} & \$0.054 & 0.582 & \$0.019\\
\system & \llamathree & \gpt & 0.662 & \$0.052 & 0.726 & {\$0.079} & 0.703 & \$0.057 & 0.558 & \$0.020\\
\system & \qwenthree & \gpt & 0.676 & {\$0.039} & 0.783 & \$0.059 & 0.645 & {\$0.043} & \textbf{0.600} & \$0.015\\
\hline
\end{tabular}

\caption{\textbf{Accuracy and cost of local-remote systems.}  Evaluation of cost and accuracy on 3 evaluations datasets. The table compares two edge-remote protocols—\naive (Section~\ref{sec:naive}) and \system (Section~\ref{sec:methods})—against edge-only and remote-only baselines.  We assess 3 local models and 1 remote model. Cost (USD) is the average per-query expense, based on \gpt rates (Jan 2025: \$2.50M/input tokens, \$10.00M/output tokens). Local model execution is assumed free (see Section~\ref{sec:prelim-setup} for cost details).
}
\label{table:main-tradeoff}

\end{table*}


\vspace{-0.5em}
\subsection{Protocol hyper-parameters}
\system has three hyper-parameters: choice of $\remotelm$ and $\locallm$ (model choice), job preparation strategy (scale of parallel workloads on-device), and  looping strategy (sequential communication protocol).

\textbf{Model choice}. Different model sizes (\textit{e.g.} 3B vs. 8B), families (\textit{e.g.} \qwen vs. \llama), and generations (\textit{e.g.} 3.1 vs. 3.2) can be used for both the $\locallm$ and the $\remotelm$. 

\textbf{Scale of parallel workload on-device}. \system has three knobs for increasing the degree of task decomposition and thus, workload parallelization: (1) number of tasks per round (\textit{i.e.} ``Extract the ARR for Q1 of 2014''), (2) number of samples per task (\textit{i.e.} number of generations created with $\locallm$, $\geq 1$), and (3) chunk size (\textit{i.e.} chunk by page, chunk by paragraph, etc; smaller chunks will send more information to cloud). These parameters are configured by $\remotelm$.

\textbf{Sequential communication protocol}. 
In practice, it is important to cap the number of times \system can loop. 
After the maximum number of rounds, the synthesis prompt is modified to force the model to produce a final answer. The choice of this maximum affects accuracy and cost. The strategy for maintaining context between rounds (simple retries vs. scratchpads) is another important hyperparameter.

\textit{We analyze these hyperparameters in \Cref{sec:results}.}

\vspace{-1em}
\section{Results}
\label{sec:results}
\begin{figure}[t]
    \centering
    \includegraphics[width=1\linewidth]{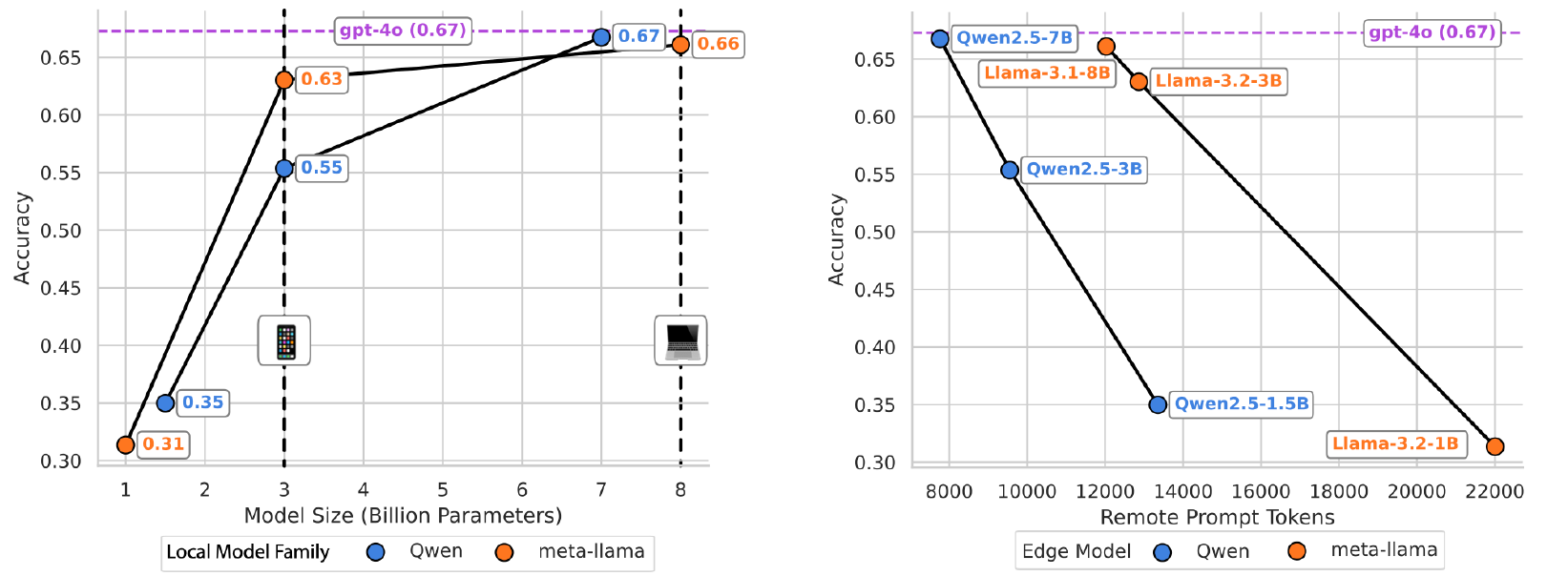}
    \vspace{-1em}
    \caption{\textbf{Trade-offs in edge model performance, communication efficiency, and cost of sequential communication.} \textbf{(Left)} Accuracy vs. local model size, with the purple dashed line showing the \gpt model baseline. \textbf{(Right)} Communication efficiency of \system with different local LMs, where larger models (7–8B) are more token-efficient.}
    \label{fig:scaling-edge-model}
  \end{figure}

Here, we analyze how the design of \system affects cost and quality. Our main takeaways are: 
\begin{itemize}
\item On average across three datasets, \system can recover $97.9\%$ of the performance of remote-only systems while spending $5.7\times$ less; 
\item We identify protocol hyper-parameters that let us flexibly trade-off cost and quality;
\item As local models grow stronger, \system becomes increasingly cost-effective.
\end{itemize}

We structure our analysis around three core design choices:
\begin{enumerate}
    \item \textbf{Model choice} \textit{How does the choice of local and remote model effect cost and quality?} We examine different model types and sizes for $\locallm$ and $\remotelm$ in \Cref{subsec:results-model}.\vspace{-0.5em}
    \item \textbf{Scaling parallel workloads on-device} \textit{How should we structure parallel workloads on the local device to maximize performance and minimize cost?} We highlight how scaling the local workloads can improve performance (\Cref{subsec:results-workloads}) and study the effects on cost.\vspace{-0.5em}
    \item \textbf{Sequential communication protocol} \textit{Can multiple rounds of communication improve quality? At what cost?} We explore this trade-off in \Cref{subsec:results-communication}.

\end{enumerate}

Our findings are detailed in Sections~\ref{subsec:results-model},~\ref{subsec:results-workloads}, and ~\ref{subsec:results-communication}. Finally, in \Cref{subsec:rag} we discuss the relationship between retrieval augemented generation and local-remote compute. 

\vspace{-0.5em}
\subsection{Experimental setup} 
\label{subsec:exp-setup}
\paragraph{Datasets and models} We evaluate \system on three benchmarks that are well suited for data-intensive reasoning: \finance, \longhealth, and \qasper. \finance tests financial document understanding with complex reasoning over reports. \longhealth focuses on tracking and interpreting longitudinal health records. \qasper assesses question answering over dense scientific papers. See Appendix~\ref{app:experiments-dataset} for details. We use two open-source model families (\llama, \qwen) as $\locallm$ and \gpt as $\remotelm$ (details in \Cref{app:experiments-models}).

\vspace{-0.5em}\subsection{Model choice}
\label{subsec:results-model}

This section explores the model requirements and generalization capabilities of \system, examining the local model sizes necessary for effective collaboration, the sensitivity of the communication protocol across different local-remote model pairings, and the longitudinal evolution of \system’ performance with advances in model capabilities over time.

\vspace{-0.5em}\paragraph{\textit{What size does $\locallm$ have to be in order to be effective in \system?}}

Our results demonstrate that \system starts being competitive with $\remotelm$-only baseline at the $3$B parameter model scale. When considering both the \qwen and \llama model families running locally, at $1$B scale, \system recovers 49.5\% of the \gpt-only baseline performance, 3B scale recovers 93.4\% and 8B recovers 97.9\% accuracy (see Table~\ref{table:main-tradeoff} for more details).

\vspace{-0.75em}\paragraph{\textit{How does the capacity of $\locallm$ affect the cost-accuracy tradeoff?}}
In our system, $\locallm$ implicitly acts as an information encoder, optimizing the Information Bottleneck objective~\citep{tishby2000information} by compressing input context while preserving predictive information (see Appendix~\ref{app:info_bottleneck}). To measure this, we analyze the tradeoff between remote ``prefill'' tokens (fewer tokens indicate greater compression) and accuracy (higher accuracy means better retention). Figure~\ref{fig:scaling-edge-model} shows that as $\locallm$ size increases, representations become more compressed and accurate, improving Information Bottleneck values. Larger $\locallm$ models trade local FLOPs for communication, with 7–8B models being 1.53× more token-efficient than 1B models. Additionally, the \qwen family follows a different tradeoff than \llama, yielding more compressed representations. This suggests that as small LMs improve, local-remote systems will become increasingly cost-efficient.


\vspace{-0.75em}\paragraph{\textit{Is \system sensitive to different local/remote pairs?}} 
We ask whether the communication protocol in \system is invariant to changing the model types (\textit{i.e.} \llama vs \qwen locally and \llama vs \gpt remotely). Our results indicate that \system performs similarly with different local-remote LM combinations (see the Table~\ref{table:main-tradeoff}): varying the $\locallm$ from \qwen to \llamathreetwo, results in performances within $\pm$ .05 performance points (see Table~\ref{table:main-tradeoff}). Furthermore, we find that holding the $\locallm$ fixed as \llamathreetwo-3B and varying $\remotelm$ from \gpt to \llama-3.3-70B leads to similar overall performances within $\pm$ 0.07 points (see Table~\ref{tab:remote-model-variations} in Appendix).

\begin{figure*}[t]
    \centering
    \includegraphics[width=\linewidth]{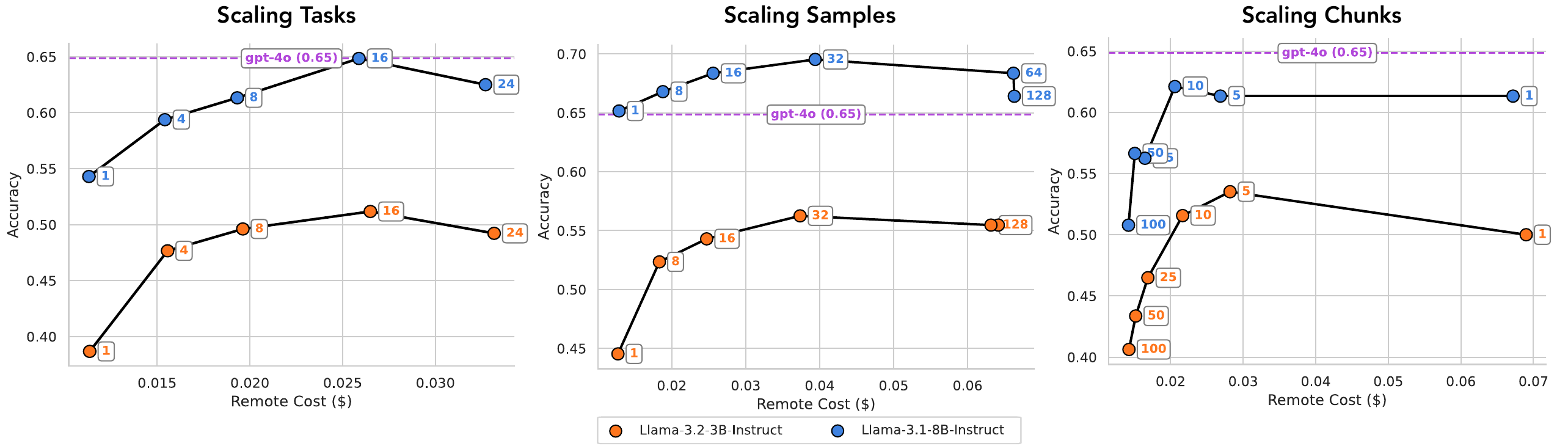}
    \caption{\textbf{Scaling parallel jobs on-device improves quality.} The x-axis represents tokens processed by the \textit{remote model}, and the y-axis shows macro-average accuracy across \longhealth and \qasper. The cloud model is \gpt. Each plot varies a different \system hyperparameter affecting parallelism, with annotated values. \textbf{(Left)} Varying the number of unique \textit{instructions}. \textbf{(Middle)} Varying the number of unique \textit{samples}. \textbf{(Right)} Varying the chunking granularity in code $\mathbf{f}$. See Section~\ref{sec:methods} for details.}
    \label{fig:scaling-samples}
\end{figure*}

\vspace{-0.75em}\paragraph{\textit{How have local / remote model capabilities changed over time, and what effects do they have on \system?}}
In Table~\ref{tab:system-snapshot}, we provide a retrospective analysis demonstrating how the quality of \system would have changed with model releases over time. 
From 2023 to 2025, the average performance of \system with the best models available has improved from 0.26 to 0.66 (see Table~\ref{tab:system-snapshot} in Appendix). Interestingly, it was only in July 2024 --- with the release of \textsc{gpt4-turbo} and \llamathreeone-8B --- that \system could have come within 12\% of the best frontier model performance at the time (see Table~\ref{tab:system-snapshot} in Appendix).

\vspace{-0.5em}\subsection{Scaling parallel workloads on-device}
\label{subsec:results-workloads}


In \system, there are three levers for maximizing local compute resources through parallelized, batched processing: (1) number of tasks per round, (2) number of samples taken per task, and (3) number of chunks. We ablate each, showing their impact on performance. We find that (1) and (3) are more cost effective ways of increasing performance.


\vspace{-0.75em}\paragraph{\textit{How does the number of tasks per round affect performance?}} Increasing tasks per round proxies task decomposition, with more sub-tasks enhancing decomposition. Raising tasks from 1 to 16 boosts performance by up to 14 points but doubles $\remotelm$ prefill costs. Optimal task count varies by query and model, but exceeding 16 reduces performance.

\vspace{-0.75em}\paragraph{\textit{How does scaling local samples affect performance?}} We explore whether increased sampling at an individual \{task, context\} level improves performance. Increased sampling enables us to better utilize the available compute resources while improving task-level accuracy~\citep{brown2024large}. Our results indicate that increasing the number samples from $1$ to $32$ can improve performance on average $7.4$ points, but comes at the cost of $5\times$ the $\remotelm$ prefill costs. This being said, increasing sampling beyond $16$ starts hurting task performance as the noise across samples is too large for the remote model to effectively distill the correct answer~\citep{kuratov2024babilong}.

\vspace{-0.75em}\paragraph{\textit{What effect does chunk size have on downstream performance?}} We test whether increasing local utilization by using more chunks per task improves performance. Our results indicate that increasing \# of chunks per task (by decreasing the number of ``pages'' per chunk from $100$ to $5$) leads to an $11.7$ point accuracy lift. However, this lift comes with a $2.41\times$ increase in $\remotelm$ prefill costs.

\vspace{-0.5em}\subsection{Scaling sequential communication}
\label{subsec:results-communication}

\begin{wrapfigure}{r}{0.5\linewidth}
    \centering
    \includegraphics[width=\linewidth]{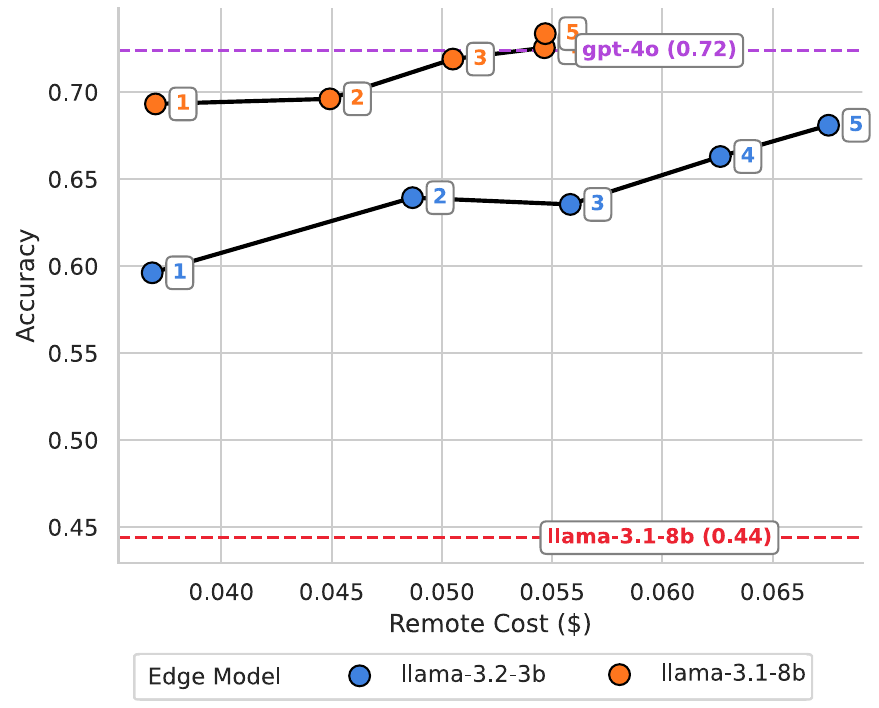}
    \caption{\textbf{Exploring the trade-off between cost and quality through multiple rounds.} The x-axis represents the remote model’s token cost, while the y-axis shows accuracy. Point labels indicate communication rounds. The purple dashed line marks \gpt’s performance as a benchmark.}
    \label{fig:scaling-rounds}
\end{wrapfigure}

Both the \naive and \system communication protocols feature \textit{sequential} communication: they allow for multiple rounds of exchange between the local and remote models.

\vspace{-0.75em}\paragraph{\textit{Does performance improve as we increase the maximum number of rounds? At what cost?}} We vary the maximum communication rounds and find it is correlated with accuracy and cost (see \cref{fig:scaling-edge-model}). By simply increasing the maximum number of rounds in \naive from $1$ to $5$, we enable a $8.5$-point lift in average accuracy across the three tasks (with \llamathreetwo on-device).
However, this accuracy improvement comes at a cost: each additional round of communication increases the cost by \$$0.006$ per query while boosting accuracy by $4.2$ points.


\vspace{-1em}\paragraph{\textit{How should we maintain context between \system rounds?}}
We experiment with two sequential protocol strategies: (1) simple retries and (2) scratchpad. 
See Section~\ref{sec:methods} for details of these strategies.
As shown in \Cref{fig:round-strategy}, both strategies show consistent increases in both accuracy and cost when increasing the maximum number of rounds, with the scratchpad strategy achieving a slightly better cost-accuracy tradeoff. 
Notably, each additional round of communication with the scratchpad strategy leads to a larger improvement in accuracy ($6.1$ accuracy points) which are mostly offset by larger increases in cost ($8.6$ dollars).



\vspace{-0.5em}\subsection{Retrieval-Augmented Generation in the Context of Local-Remote Compute}
\label{subsec:rag}

We further investigate the interplay between local-remote compute paradigms (\textit{e.g.}, \system) and retrieval-augmented generation (RAG), analyzing their complementary strengths and trade-offs in data-intensive reasoning tasks. Through empirical evaluations, we examine these methods on financial document extraction (\finance) and long-document summarization (\textsc{Booookscore}). \textit{See Appendix ~\ref{app:rag} for a more detailed treatment.}

\subsubsection{Comparison of \system and RAG on \finance}

RAG performs well for financial document analysis, where relevant information is found in specific sections. In \Cref{fig:rag-comparison} (left), we compare \system, \naive, and RAG using BM25 and OpenAI's \texttt{text-embedding-3-small} embeddings~\citep{article,neelakantan2022text}. A chunk size of 1000 characters balances retrieval accuracy and efficiency (\Cref{fig:rag-comparison} (center)).

Adjusting the number of retrieved chunks allows RAG to optimize quality and cost. When BM25-based RAG retrieves 50+ chunks, it surpasses the performance of a fully context-fed remote model. However, RAG does not match the cost-effectiveness of \naive. When compared to \system, RAG using OpenAI embeddings achieves similar cost-quality trade-offs but may overlook nuanced financial signals across sections.

\subsubsection{Comparison of \system and RAG on Summarization Tasks}

Unlike \finance, RAG struggles with summarization due to its reliance on retrieval. Unlike the financial parsing tasks, summarization requires reasoning over information \textit{dispersed} across the document.  We evaluate \system, RAG (with BM25 andEmbedding retrievals), and a \gpt-only baseline on \textsc{BooookScore}~\citep{chang2023booookscore}, a long novel summarization dataset.

\paragraph{Evaluation}
\begin{itemize}
\item \textbf{Qualitative Analysis:} As shown in App. Table~\ref{tab:story-summaries}, \system generates summaries with richer entity mentions and more coherent narratives than RAG. It is also $9.3\times$ more token-efficient than \gpt-only (11,500 vs. 108,185 tokens).
\item \textbf{Quantitative Analysis:} Using \textsc{claude-3.5-sonnet} for evaluation, \system achieves near-parity with \gpt-only, while RAG-based methods underperform (Table~\ref{tab:summary_qualitative_scores} in Appendix).
\end{itemize}

These results highlight RAG's effectiveness in structured tasks like \finance but its shortcomings in long-form synthesis which requires reasoning over information dispersed across the document. 



\vspace{-1em}\section{Discussion} 
As local models continue to improve, studying \system will provide valuable insights into evolving workload distribution and the growing role of local compute.
As local models continue to improve, systems like \system will become increasingly valuable across a range of important workloads.

In this work, we explore two protocols -- \naive and \system -- for collaboration between on-device and cloud LMs. Our results demonstrate that it is possible to reduce the cost of cloud computing workloads $5-26\times$ by enabling remote LMs to effectively communicate with local LMs and delegate work to them. We explore the broader implications as well as the research opportunities this approach unlocks.


\vspace{-0.5em}\paragraph{User experience} Soon, commodity hardware—laptops, smartphones, and IoT devices—will feature powerful GPUs, enabling always-on local models for complex tasks like code refactoring, document analysis, and retrieval. Additionally, this advancement is expected to reduce users' reliance on API-based cloud LMs, leading to lower operational costs.

\vspace{-0.5em}\paragraph{Local-remote model co-design} \system demonstrates the promise of ``collaboration'' between local and remote LMs. Future work could advance this approach in two directions. First, the LMs we investigated were trained independently and therefore might not know each others capabilities and limitations. Communication can be made more efficient by training these models for collaboration. Second, model co-design could enable going beyond natural language as the modality of communication; models can exchange more compressed real-valued representations.


\vspace{-0.5em}\paragraph{Improvement of \system over time} Our analysis in Section~\ref{app:model_history} highlights the rapid advancements in local model capabilities over the past 15 months. As local-remote capabilities continue to evolve, studying \system will provide valuable insights into shifts in workload distribution and the increasing utilization of local compute resources.

\vspace{-1em}\section{Acknowledgements} 
We thank Michael Zhang, Jordan Jurafsky, Mayee Chen, Jon Saad-Falcon, Ben Viggiano, Michael Wornow, Aditi Jha, Ben Spector and Neel Guha for their helpful feedback and discussion during this work. We
thank the Hazy Research lab and Together AI for supporting this work. 

We gratefully acknowledge the support of NIH under No. U54EB020405 (Mobilize), NSF under Nos. CCF2247015 (Hardware-Aware), CCF1763315 (Beyond Sparsity), CCF1563078 (Volume to Velocity), and 1937301 (RTML); US DEVCOM ARL under Nos. W911NF-23-2-0184 (Long-context) and W911NF-21-2-0251 (Interactive Human-AI Teaming); ONR under Nos. N000142312633 (Deep Signal Processing); Stanford HAI under No. 247183; NXP, Xilinx, LETI-CEA, Intel, IBM, Microsoft, NEC, Toshiba, TSMC, ARM, Hitachi, BASF, Accenture, Ericsson, Qualcomm, Analog Devices, Google Cloud, Salesforce, Total, the HAI-GCP Cloud Credits for Research program,  the Stanford Data Science Initiative (SDSI), AN is supported by the Knight-Hennessy Fellowship and NSF GRFP and members of the Stanford DAWN project: Meta, Google, and VMWare. SL is supported in part by grants from the NIH (R01NS131987, U01NS136507, R01NS130789, 
 RF1MH133778) and NSF (2223827) as well as fellowships from the Sloan, Simons, and McKnight Foundation. The U.S. Government is authorized to reproduce and distribute reprints for Governmental purposes notwithstanding any copyright notation thereon. Any opinions, findings, and conclusions or recommendations expressed in this material are those of the authors and do not necessarily reflect the views, policies, or endorsements, either expressed or implied, of NIH, ONR, or the U.S. Government.   


\bibliography{iclr2025_conference}
\bibliographystyle{iclr2025_conference}

\appendix
\section{Extended Related Work}
\label{app:related-work}


\paragraph{Orchestration of LMs }
Recent works attempt to improve long document processing by taking a divide-and-conquer approach akin to \system. Instead of using single LM calls with the entire context, the task is decomposed into smaller tasks to be executed on chunks of context. \citep{zhang2024chain, zhou2024llm} use a predefined protocol for chunk processing (defined by a prompt). \citep{shankar2024docetl} performs a more involved automated pipeline optimization (via agent-based rewrite directives). Crucially, none of the works study the cost-efficient interaction between a small local LM and large remote LM and instead focus exclusively on larger LMs (70B parameters and above). Moreover, they do not explore multi-round communication patterns for document analysis. 

\paragraph{Long-context management techniques} These works aim to improve \emph{single LM} accuracy in long context tasks. \citep{russak2024writing} prefill the context using \emph{chunks} of the document, summarize each chunk (using a predefined prompt), and aggregate the results. This improves accuracy and requires marginal additional computation. PRISM similarly \citep{jayalath2024long} processes the context as a stream of chunks, and writes important information into a typed data structure which can be amended as needed. MemGPT \citep{packer2023memgpt} proposes a virtual memory paging system inspired by operating systems, where the LLM manages information across main context (akin to RAM) and external storage (akin to disk). When approaching context limits, the system actively decides what information to preserve and can later retrieve this information through paginated function calls. Orthogonally, other methods explore the usage of code for context management~\citep{arora2023evaporate}.



\paragraph{Cost-efficient multi-LLM Systems} 
A plethora of recent works show that multiple LMs can collaborate on a task to improve both accuracy and efficiency \citep{guo2024large}. The most similar work is perhaps \citep{wang2024mixture} which neither investigates LMs with with asymmetric capabilities nor optimizes for local compute efficiency. 


\paragraph{Model routing techniques} Our work studies a collaboration of LMs, and thus should be differentiated from model routing techniques \citep{chen2024more,chen2023frugalgpt} that route a prompt to the appropriate single LM that can completely answer it using the full context. This is often done for cost reduction, identifying that simple tasks can be executed by smaller and cheaper LMs. 

\paragraph{Compound LM systems} Recent works explore the use of LMs as part of more elaborate pipelines that, retrieval models, tool use, and more. \citep{saad2024archon,khattab2023dspy,yuksekgonul2024textgrad} seeks to optimize the pipeline architecture and prompts using different approaches, which we do not pursue on this work.

\paragraph{Retrieval-Augmented Generation (RAG)} RAG is a hybrid approach that integrates information retrieval into the text generation process, leveraging external knowledge sources to enhance the output of language models (LMs). Instead of relying solely on parametric memory, RAG reduces the number of tokens processed by an LM by first retrieving a subset of relevant documents or document chunks and appending them as context to the LM~\citep{lewis2020retrieval, karpukhin2020dense, lee2019latent, izacard2021unsupervised, guu2020retrieval}. This retrieval step mitigates issues such as hallucination and knowledge staleness, which are common in traditional autoregressive models~\citep{shuster2021retrieval, petroni2019language}. We differ in two ways: first, our local LM can perform tasks beyond information extraction, such as summarization or reasoning. Second, by performing arbitrary tasks on document chunks, the small LM communicates its compact answer instead of the raw document chunk, which amounts to sending fewer tokens to remote.

\paragraph{Speculative decoding}
Speculative decoding \citep{leviathan2023fast,zhang2024fast,chen2024sequoia} techniques are addressing the different question of how to effectively sample from the distribution of a large LM by only sampling from smaller LM and using the large LM for cheaper, likelihood evaluation (using the ``acceptance-complement algorithm'' \citep{devroye2006nonuniform}). It neither considers a collaboration between two LMs, nor attempts to minimize the communication between them.


\paragraph{On-device language models for privacy}
~\citet{siyan2024papillon, zhang2024cogenesis} attempt to prevent leaks of private information to a cloud-hosted LM API by mediating the communication with a local privacy-aware LM that removes private information from the prompt. While the local-remote LM setup bears resemblance to ours, we do not study the aspects of privacy, but rather focus on reducing cloud costs by delegating work to devices while maintaining accuracy. Moreover, we have additional focus on local runtime efficiency.

\paragraph{Local-remote systems} Recent work has explored efficient routing patterns between local and remote computation for LM workloads, albeit without two models communicating or collaborating on a solution.
\citep{jin2024collm} partition a single LLM with early layers on the edge and later layers in the cloud, routing to the cloud when confidence is low. \citep{yang2024perllm}
propose a complementary task scheduling framework that routes to cloud or local based on resource constraints and service requirements.

\section{Extended Description of Experimental Setup}
\label{app:experiments}
\subsubsection{Dataset Details}
\label{app:experiments-dataset}

In this section we provide additional details on dataset preparation. In order to extend the context length of the problems in \longhealth and \qasper, we make a few modification to the dataset.

\textbf{\finance} We filter the original \finance to include only the numerical reasoning, resulting in a dataset of length $64$. Each sample has an average context length of $142.9K (\pm 79224.32$).

\textbf{\longhealth}
In the original instantiation of the \longhealth dataset, each question is paired with a set of medical documents corresponding to a single patient. To increase the complexity of the dataset, we include medical documents from $10$ other patients in the context. We evaluate over the entire dataset ($400$ problems) for results reported in Table~\ref{table:main-tradeoff}. Each sample has an average context length of $120.1K (\pm 1,237)$ tokens. For all ablations in Section~\ref{sec:results}, we use a fixed subset of $128$ problems. 

\textbf{\qasper} Similarly, in the \qasper dataset, the original dataset provides questions that are associated with a single scientific paper. In order to increase complexity, we include $10$ other papers in the context.  We evaluate over a random subset of $500$ problems for results reported in Table~\ref{table:main-tradeoff}. Each sample has an average context length of $54281$ tokens ($\pm 2403)$. For all ablations in Section~\ref{sec:results}, we use a fixed subset of $128$ problems.

\subsubsection{Model Details}
\label{app:experiments-models}
\textbf{Local Models}. For \qwen we use the following models: \qwen-1.5-Instruct, \qwen-3B-Instruct, \qwen-7B-Instruct. For \llama, we use the following models: \llamathreetwo-1B-Instruct, \llamathreetwo-3B-Instruct, \llama-3.1-8B-Instruct.

\textbf{Remote Models}. We use \gpt and \llamathreetwo-70B-Instruct, \llamathreeone-70B-Instruct

All ``local-only'' and ``remote-only'' experiments are run with temperature of $0.2$. For all \system experiments run in Table~\ref{table:main-tradeoff}, we run the $\remotelm$ with a temperature of $0.0$ and $\locallm$ with a temperature of $0.2$ for \finance and $0.00001$ for \qasper and \longhealth.

\section{Extended Discussion of Cost Model}
\label{app:edgecost}
Here, we explain in detail the costs of the different communication protocols discussed in this paper---remote-only, \naive, and \system---with a strong focus on the latency of these methods.
This section is organized as follows:
\begin{itemize}
    \item Section~\ref{app:edgecost_background}: We review background on language model inference, to motivate our cost and latency models.
    \item Section~\ref{app:edgecost_models}: We present mathematical models for the latency of the remote-only, \naive, and \system protocols.
    \item Section~\ref{app:edgecost_bound}: We present Proposition~\ref{prop:latency_bound}, an upper bound on the total latency of \system, relative to that of the remote-only model, demonstrating that \system is not much slower than the naive approach of performing the full query in the cloud.
    As an example, we show that a Llama-8B model on a GTX-4090 GPU collaborating via \system with a Llama-405B model on a $8\times$H100 server is at most $4.75\times$ slower than the remote-only protocol.
\end{itemize}

\subsection{Background on language model inference}
\label{app:edgecost_background}
Language model inference consists of a sequence of forward passes through a model, one for prefill (\textit{i.e.} input) followed by one for each additional token generated (\textit{i.e.} output). 
At low/medium batch sizes, each forward pass after prefill is I/O bound, meaning the time it takes to load weights from memory exceeds the time it takes to actually compute the output. 
As the batch size increases, the computational cost of the forward pass eventually exceeds the I/O cost. 
Strikingly, for most models and hardware, this happens at a batch size $>100$~\citep{leviathan2023fast,chen2024sequoia}.
As a result of this transition from being I/O bound to being compute bound, we can model (as is common in the literature) the cost of running a forward pass as a piecewise linear function $C_{\mathcal{M}, \mathcal{E}}(n) = \max(\lambda, \alpha \cdot n + \beta)$ of the number of tokens $n$ being processed.
This is because for small $n$, the IO cost dominates (and is roughly constant as $n$ grows), whereas at larger $n$ the compute cost dominates and scales roughly linearly with $n$ (assuming $n$ is not \textit{too} large).

In the cloud, the provider can batch generation requests from multiple users to keep hardware utilization high.
Therefore, the cost of each output token is typically within a small multiple of the cost of each input token, and the total cost of processing the request scales as $n_{prefill} + \alpha \cdot n_{decode}$, for some small $\alpha \leq 5$. 

On-device, we cannot assume we'll have enough concurrent user requests to form a large enough batch to achieve high utilization. As a result, the latency of a request does not scale linearly with the number of tokens. A single request can occur similar latency to hundreds run in parallel. As a result, tokens are a poor proxy for cost on-device and we instead measure latency in micro experiments (see \Cref{subsec:results-workloads}). 

\subsection{Latency models for all protocols: Remote-only, \naive, \system}
\label{app:edgecost_models}
We now model the latency of each of these protocols (remote-only, \naive, \system).
We will then use these results in the following section to upper bound the latency of \system by a scalar multiple of the latency of the remote-only protocol.

\textbf{First, we introduce the following assumptions and notation}:
\begin{itemize}
\item We assume we have a local GPU (\textit{e.g.} RTX-4090) with peak compute $F_l$ (flops/sec), and peak bandwidth $M_l$ (bytes/sec), and a remote GPU (\textit{e.g.} H100) with peak compute $F_r$ (flops/sec), and peak bandwidth $M_r$ (bytes/sec), 
\item We also assume for now simple transformer architectures for both the local and remote models:
\begin{itemize}
\item $\locallm$: $L_l$ layers, each with $8d_l^2$ params in MLP (Up/down projections each of size $d_l\times 4d_l$, and $4d_l^2$ parameters in the $W_{Q,K,V,O}$ projections.
The total memory required for the (non-embedding/LM head) parameters is thus $P_l = 2 \cdot 12 L_l d_l^2$.
For simplicity, we assume the memory for the LM head is small relative to $P_l$.
\item $\remotelm$: Equivalent architecture to the $\locallm$, but with $L_r$ layers, $d_r$ hidden dimension, and $P_r$ total non-embedding/LM-head parameter memory (again assumed to be much greater than the number of LM head parameters).
\end{itemize}
\item We model the number of input/output tokens of each protocol as follows, letting $n$ denote the number of tokens in the original document:
\begin{itemize}
    \item \textbf{Remote-only}: $n$ prefill tokens and $n_{out}^r$ decode tokens.
    Note that we assume---here and below---that the number of tokens in the query is negligible relative to $n$.
    We assume $n \gg n_{out}^r$ so we can effectively ignore the KV-cache load time for the output tokens.
    \item \textbf{\naive}: For $\locallm$, we assume $n$ prefill tokens and $n_{out}^l$ decode tokens.
    For $\remotelm$, we assume $n_{out}^l$ prefill tokens, and $n_{out}^r$ decode tokens.
    In the case of multiple rounds of communication, the KV cache for the document can be stored to avoid recomputation.
    \item \textbf{\system}: For $\locallm$, we assume $n/c$ prefill tokens per chunk ($c$ chunks total), and $n_{out}^l$ decode tokens per job (though we assume only $p$ fraction of output jobs do not abstain).
    For $\remotelm$, we assume $J \cdot n_{out}^l \cdot p$ prefill tokens, and $n_{out}^r$ decode tokens, letting $J=cks$ denote the total number of jobs in \system ($c$ chunks, $k$ instructions, $s$ samples).
    In the case of multiple rounds of communication, the KV cache for each document chunk can be stored to avoid recomputation.
\end{itemize}
\item Throughout, we use the fact that a $[m \times n] \cdot [n \times k]$ matmul takes $2\cdot mnk$ flops, and assume model parameters are stored in half-precision (2 bytes/param).
\end{itemize}


We are now ready to present the latency models for the three protocols (remote-only, \naive, \system).

\subsubsection{Remote-only}
\begin{itemize}
    \item \textbf{Prefill}: We are compute bound, so time is approximately given by $total\_flops / F_r$.
    We can break down $total\_flops$ into the matmuls (MLP up/down projections, and QKVO operations) and attention operations.
    \begin{itemize}
        \item \textbf{Matmuls}: $2 \cdot 12 n d^2$ per layer. 
        Equivalent to a $[n \times d_r] \cdot [d_r \times 12d_r]$ matmul.
        \item \textbf{Attention}: $2 \cdot n^2 d_r$ per layer.
        Equivalent to $[n \times d_r] \cdot [d_r \times n]$ matmul.
        \item \textbf{Time}: $L_r \cdot (24 n d_r^2 + 2 n^2 d_r) / F_r = (nP_r + 2 L_r d_r n^2) / F_r$.
    \end{itemize}
    \item \textbf{Decode}: We are memory bound (batch size 1 for Minion), so time is approximately given by $total\_memory / M_r$ per decode step.
    We can break down $total\_memory$ into model parameters and KV cache.
    \begin{itemize}
        \item \textbf{Model parameters}: $2 \cdot 12 d_r^2$ bytes per layer.
        \item \textbf{KV-cache}: $2 \cdot 2 n d_r$ bytes per layer (K and V are each  $[n \times d]$ matrices).
        \item \textbf{Time}: $L_r \cdot n_{out}^r \cdot (24 d_r^2 + 4 n d_r) / M_r =  n_{out}^r (P_r + 4 L_r d_r n) / M_r$.
    \end{itemize}
\end{itemize}
\noindent \textit{Total time} is given by the sum of prefill and decode times: 
    $$T_{remote} = \frac{nP_r + 2 L_r d_r n^2}{F_r} + \frac{n_{out}^r (P_r + 4 L_r d_r n )}{M_r}$$

\subsubsection{\naive}
The latency of the $\locallm$ in the \naive protocol can be modeled equivalently to the latency of the remote-only protocol, but replacing the remote parameters with the corresponding local ones.
Thus, total local latency is:
    $$T_{local}^{\naive} = \frac{nP_l + 2 L_l d_l n^2}{F_l} + \frac{n_{out}^l (P_l + 4 L_l d_l n )}{M_l}$$

The total remote latency can also be expressed using these same equations, but with $n_{out}^l$ prefill tokens, and $n_{out}^r$ decode tokens.
$$T_{remote}^{\naive} = \frac{n_{out}^l P_r + 2 L_r d_r (n_{out}^l)^2}{F_r} + \frac{n_{out}^r (P_r + 4 L_r d_r n_{out}^l )}{M_r}$$

\subsubsection{\system}
The $\locallm$ latency of the \system protocol has some important differences from the \naive protocol---the prefill computation avoids cross-chunk attention (which saves time), while the decode operations can actually be compute bound if batching of the different jobs is done.
We review these details below:

\begin{itemize}
    \item \textbf{Prefill}: We are compute bound, so time is approximately given by $total\_flops / F$. We can break down $total\_flops$ into the matmuls (MLP up/down projections, and QKVO operations) and attention operations.
    \begin{itemize}
        \item \textbf{Matmuls}: $2 \cdot 12 n d_l^2$ per layer. Equivalent to $c$ $[n_c \times d_l] \cdot [d_l \times 12d_l]$ matmuls (where $n_c = n/c$).
        \item \textbf{Attention}: $2 \cdot c n_c^2 d_l = 2 \cdot c \; (n/c)^2 d_l = 2n^2 d_l / c$ per layer.  Equivalent to $c$ $[n_c \times d_l] \cdot [d_l \times n_c]$ matmuls.
        \item \textbf{Time}: $L_l \cdot (24 n d_l^2 + 2 n^2 d_l / c) / F = (n P_l + 2 L_l d_l n^2 /c) / F$.
    \end{itemize}
    \item \textbf{Decode}: We will now assume we are \textbf{compute bound during decode}, because we have many jobs ($ks$) per chunk, and many chunks ($c$) per document, which we can batch together. Thus, time is approximately given by $total\_flops / F_l$ per decode step. We can break down $total\_flops$ into matmuls and attention. The flops below are per job, per output token (so for total flops we will multiply by $n_{out}^l \cdot pcks$):
    \begin{itemize}
        \item \textbf{Matmuls}: $2 \cdot 12 d_l^2$ per layer. Equivalent to a $[1 \times d_l] \cdot [d_l \times 12d_l]$ matmul.
        \item \textbf{Attention}: $2 \cdot n_c d_l = 2d_l \, n/c$ per layer. Equivalent to $[1 \times d_l] \cdot [d_l \times n_c]$ matmul.
        \item \textbf{Time}: $L_l \cdot n_{out}^l \cdot pcks \cdot (24 d_l^2 + 2 d_l n/c) / F = n_{out}^l \cdot pcks \cdot (P_l + 2 L_l d_l n/c) / F$.
    \end{itemize}
\end{itemize}
The \textit{total local latency} for \system is given by the sum of prefill and decode times: 
\systemmath
$$T_{local}^{\systemmath} = \frac{n P_l + 2 L_l d_l n^2 /c}{F_l} + \frac{n_{out}^l \cdot pcks \cdot (P_l + 2 L_l d_l n/c)}{F_l}.$$

The \textit{total remote latency} for \system can be expressed using the same equations as \naive, but with $pcks \cdot n_{out}^l$ prefill tokens, and $n_{out}^r$ decode tokens.
$$T_{remote}^{\systemmath} = \frac{(pcks \cdot n_{out}^l) P_r + 2 L_r d_r (pcks \cdot n_{out}^l)^2}{F_r} + \frac{n_{out}^r (P_r + 4 L_r d_r (pcks \cdot n_{out}^l) )}{M_r}$$

\subsection{\system vs. remote-only comparison}
\label{app:edgecost_bound}

\begin{proposition}
\label{prop:latency_bound}
Assume $n_{out}^l \cdot pcks = an$, for some $a < 1$, and that $F_{r,l}$, $d_{r,l}$, and $L_{r,l}$ are all as defined in Appendix~\ref{app:edgecost_models}. In this case, we can show that the ratio of total latency of \system vs. the remote-only protocol is upper-bounded by the following expression:

\begin{eqnarray*}
\frac{T_{remote}^{\systemmath} + T_{local}^{\systemmath}}{T_{remote}} &<& 1 + \big(1 + a \big) \cdot \frac{F_r}{F_l} \cdot \frac{L_l d_l}{L_r d_r}. \\
\end{eqnarray*}

\begin{proof}
    
Let's assume $n_{out}^l \cdot pcks = an$, for some $a < 1$.

\begin{eqnarray*}
T_{local}^{\systemmath} &=& \frac{n P_l + 2 L_l d_l n^2 /c}{F_l} + \frac{an \cdot (P_l + 2 L_l d_l n/c)}{F_l} \\
&<& \big(1 + a\big) \cdot \frac{n P_l + 2 L_l d_l n^2 /c}{F_l} \\
T_{remote}^{\systemmath} &=& \frac{(an) P_r + 2 L_r d_r (an)^2}{F_r} + \frac{n_{out}^r (P_r + 4 L_r d_r (an) )}{M_r} \\
&<& a \bigg(\frac{nP_r + 2 L_r d_r n^2}{F_r} + \frac{n_{out}^r 4 L_r d_r n}{M_r}\bigg) + \frac{n_{out}^r P_r}{M_r}\\
T_{remote} &=& \frac{nP_r + 2 L_r d_r n^2}{F_r} + \frac{n_{out}^r (P_r + 4 L_r d_r n )}{M_r}
\end{eqnarray*}
Thus, it is easy to see that $\frac{T_{remote}^{\systemmath}}{T_{remote}} < 1$.
Now let's look at $\frac{T_{local}^{\systemmath}}{T_{remote}}$, and show it is upper bounded by a constant:

\begin{eqnarray*}
\frac{T_{local}^{\systemmath}}{T_{remote}} &<& \frac{\big(1 + a\big) \cdot \frac{n P_l + 2 L_l d_l n^2 /c}{F_l}}{ \frac{nP_r + 2 L_r d_r n^2}{F_r}} \\
&=& \big(1 + a \big) \cdot \frac{F_r}{F_l} \cdot \frac{n P_l + 2 L_l d_l n^2 /c}{nP_r + 2 L_r d_r n^2} \\
&\leq& \big(1 + a \big) \cdot \frac{F_r}{F_l} \cdot \max\bigg(\frac{P_l}{P_r}, \frac{L_l d_l}{L_r d_r c}\bigg) \\
&=& \big(1 + a \big) \cdot \frac{F_r}{F_l} \cdot \max\bigg(\frac{L_l d_l^2}{L_r d_r^2}, \frac{L_l d_l}{L_r d_r c}\bigg) \\
&<& \big(1 + a \big) \cdot \frac{F_r}{F_l} \cdot \frac{L_l d_l}{L_r d_r}.\\
\end{eqnarray*}

Thus, combining the above two results we can see that:
\begin{eqnarray*}
\frac{T_{remote}^{\systemmath} + T_{local}^{\systemmath}}{T_{remote}} &<& 1 + \big(1 + a \big) \cdot \frac{F_r}{F_l} \cdot \frac{L_l d_l}{L_r d_r}. \\
\end{eqnarray*}

\end{proof}

\end{proposition}

\textbf{Real example}: Let's assume that the local GPU is a RTX 4090 ($F_l \approx 160$ TFLOPS), the remote server is a full node of 8 H100s ($F_r \approx 8000$ TFLOPS across full node), the local model is Llama-8B ($L_l=32$, $d_l=4096$), and the remote model is Llama-405B ($L_l=126$, $d_l=16384$).
Furthermore, let's assume $a \approx 0.2$, which is actually a bit larger than we see in practice.
In this case:
\begin{eqnarray*}
1 + \big(1 + a \big) \cdot \frac{F_r}{F_l} \cdot \frac{L_l d_l}{L_r d_r} &\approx& 1 + 1.2 \cdot \frac{8000}{160} \cdot \frac{32 \cdot 4096}{126 \cdot 16384} \\
&\approx& 1 + 1.2 \cdot 50 \cdot \frac{1}{16} \\
&=& 4.75. \\
\end{eqnarray*}

Note that if we perform multiple rounds of \system, this ratio gets multiplied by at most the number of rounds, though as mentioned previously, we can save time by only performing prefill on all the document chunks in the first round.




\section{Extended discussion of methods}
\label{app:methods}

\subsection{Extended description of \naive}
\label{app:methods-naive}

In this section, we describe \naive, a baseline local-remote communication protocol. 
We ask whether we can reduce remote prefill tokens, and thus cost, by simply orchestrating a free-form conversation between the $\locallm$ and the $\remotelm$ in which only the $\locallm$ has direct access to the context $\bc$.

The protocol proceeds with initialization step followed by a simple correspondence between the two models, which terminates when the remote model can answer the question or a maximum iteration limit is reached. 

\paragraph{Iteration $i = 1$: Initialize.}
The $\remotelm$ receives the task query $\bq$ along with a system prompt $\bp_{\text{remote}}$ that instructs it to  interact with a small LM that has full access to context. It outputs a first message $\bmm_{\text{remote}}^{(1)}$:
\begin{align*}
    \bmm_{\text{remote}}^{(1)} \sim \textcolor{blue}{\remotelm}(\bq, \bp_{\text{remote}})
\end{align*}
The message is then provided to $\locallm$, along with the full context $\bc$, the query $\bc$,
and a minimal system prompt $\bp_{\text{local}}$ that instructs it to answer questions on the context:
\begin{align*}
    \bmm_{\text{local}}^{(1)} \sim \textcolor{blue}{\locallm}(\bmm_{\text{remote}}^{(1)}, \bq, \bp_{\text{local}}, \bc)
\end{align*}
\paragraph{Iteration $i > 1$.}
\textbf{Step 1: Message from remote to local.}
$\remotelm$ consumes the conversation history and outputs new messages: 
\begin{align*}
    \bmm_{\text{remote}}^{(i)} \sim \remotelm(\bmm_{\text{remote}}^{(:i - 1)}, \bmm_{\text{local}}^{(:i - 1)} , \bq, \bp_{\text{remote}})
\end{align*}
In its message, $\remotelm$ indicates whether it has sufficient information to terminate the loop and answer the question, or alternatively raises additional questions.

\textbf{Step 2: Message from local to remote}
$\locallm$ consumes the latest remote message and conversation history, and outputs $\bmm_{\text{local}}^{(i)}$. 
\begin{align*}
    \bmm_{\text{local}}^{(i)} \sim \textcolor{blue}{\locallm}(\bmm_{\text{remote}}^{(:i - 1)}, \bmm_{\text{local}}^{(:i - 1)}, \bq, \bp_{\text{local}}, \bc)
\end{align*}
We then increment the iteration $i$ and loop back to \textbf{Step 1} until the break condition is met or we reach a maximum number of iterations.

\subsection{Information Bottleneck Perspective}\label{app:info_bottleneck}
How does local model capacity affect the cost-accuracy tradeoff?

The Information Bottleneck (IB) principle \citep{tishby2000information} provides a useful analogy.
One communication round of a local-remote system does as follows:
\begin{align*}
    \bz &\sim p(\bz|\bc) \quad \text{[Extract info. from context]} \\
    \by &\sim p(\by|\bz) \quad \text{[Predict outcome from extracted info]}
\end{align*}
The IB principle seeks to find a $p(\bz \mid \bc)$, our $\locallm$, as follows:
\begin{equation}
\min_{p(\bz \mid \bc)} \;\Bigl[\, I(C;Z) \;-\; \beta \, I(Z;Y) \Bigr],
\label{eq:ib_formulation}
\end{equation}
\textit{i.e.} find a mapping that forces the latent representation to be maximally informative of the label $I(Z;Y)$ and minimally informative of the input $I(C;Z)$, with a tradeoff parameter $\beta$. Here, we do not optimize the mapping $p(\bz \mid \bc)$ but instead only get to choose it by setting $\locallm$.

Since we cannot compute these quantities in closed form for nonlinear distributions over tokens, we use (coarse) empirical proxies as follows. As a proxy for $I(C;Z)$, we compute the number of prefill tokens sent to $\remotelm$, capturing the intuition that more tokens carry more information on the input.
$I(Z;Y)$ is estimated as the average accuracy of the local-remote system, quantifying the preservation of task-relevant information in $\bz$. While these proxies do not exactly match the underlying mutual informations, they capture the core tension of compressing the input vs.\ preserving predictive power.

We plot these quantities in Figure~\ref{fig:lm-bottleneck}. We find that  across both $\qwen$ and $\llama$ model families, as we increase $\locallm$ size, we send fewer tokens to $\remotelm$ ($\approx I(C;Z) \downarrow$), and improve accuracy ($\approx I(Z;Y) \downarrow$). We find that $\llama$ has higher $\approx I(C;Z)$ and higher $\approx I(Z;Y)$.

\section{Extended Results}

\begin{figure}[t]
    \centering
    \includegraphics[width=0.7\linewidth]{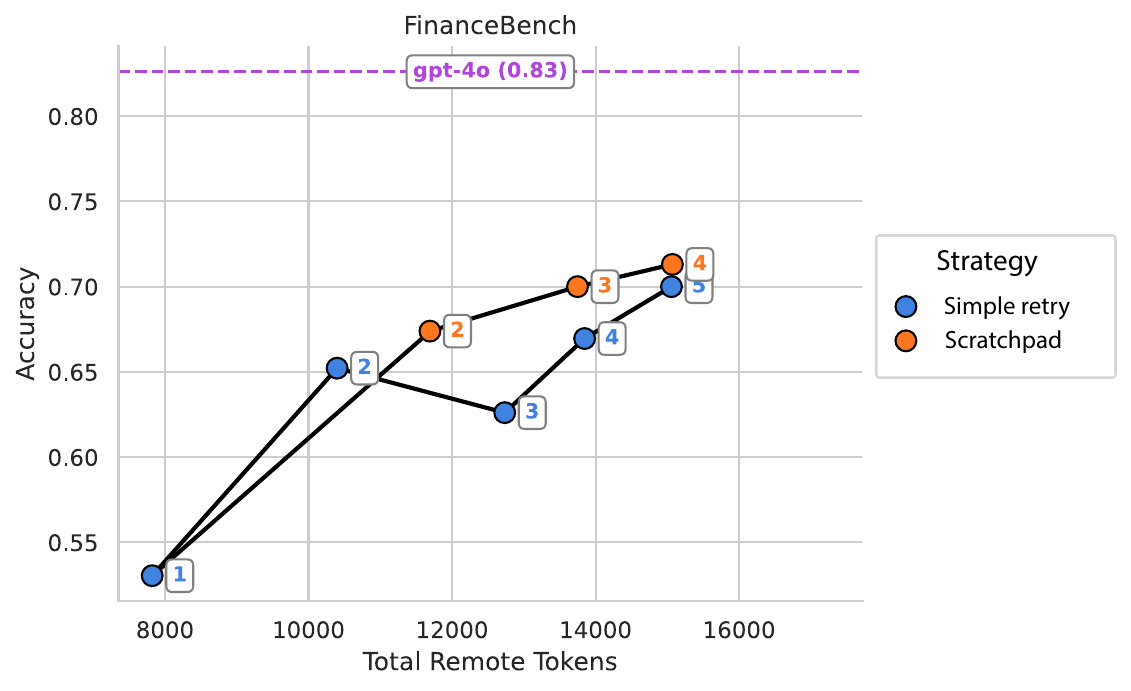}
    \caption{\textbf{Comparing strategies for maintaining context between \system rounds.} The x-axis represents the number of tokens processed by the \textit{remote model}, while the y-axis shows the accuracy achieved.}
    \label{fig:round-strategy}
\end{figure}

\subsection{Model Analysis}\label{app:model_history}
\begin{table*}[]
\centering
\scriptsize

\begin{tabular}{l l l c c c c}
\toprule
\multicolumn{1}{c}{\textbf{Local Model}} & \multicolumn{1}{c}{\textbf{Remote Model}} & \multicolumn{1}{c}{\textbf{Release Date}} & \multicolumn{1}{c}{\textbf{Accuracy (Longhealth)}} & \multicolumn{1}{c}{\textbf{Accuracy (QASPER)}} & \multicolumn{1}{c}{\textbf{Accuracy (Finance)}} \\ 
\midrule
llama-3B & gpt-4o & May 2024 & 0.7025 & 0.598 & 0.7826 \\ 
llama-3B & gpt-4-turbo & April 2024 & 0.6247 & 0.614 & 0.6304 \\ 
llama-3B & gpt-3.5-turbo-0125 & Jan  2024 & 0.2157 & 0.4314 & 0.1707 \\ 
llama-3B & gpt4o-mini & July 2024 & 0.6275 & 0.568 & 0.6522 \\ 
llama-3B & llama3-70B-Instruct-Turbo & April 2024 & 0.3525 & 0.144 & 0.1818 \\ 
llama-3B & llama3.1-70B-Instruct-Turbo & July 2024 & 0.6193 & 0.514 & 0.4348 \\ 
llama-3B & llama3.3-70B-Instruct-Turbo & December 2024 & 0.6658 & 0.534 & 0.6739 \\ 
\bottomrule
\end{tabular}
\caption{Accuracy Results for Longhealth, QASPER, and Finance across Various Models}
\label{tab:remote-model-variations}
\end{table*}
\begin{table*}[]
\centering
\scriptsize

\begin{tabular}{l l c c l}
\toprule
\multicolumn{1}{c}{\textbf{Local Model}} & \multicolumn{1}{c}{\textbf{Remote Model}} & \textbf{Accuracy (Longhealth)} & \textbf{Accuracy (QASPER)} & \multicolumn{1}{c}{\textbf{System Date}} \\ 
\midrule
Llama-2-7b-chat-hf & gpt-4-1106-preview & 0.340 & 0.178 & November 2023 \\ 
Llama-3.1-8B-Instruct & gpt-4-turbo & 0.645 & 0.528 & April 2024 \\ 
Llama-3.1-8B-Instruct & gpt-4o & 0.740 & 0.582 & July 2024 \\ 
--- & gpt-4-turbo & 0.768 & 0.391 & April 2024 \\ 
\bottomrule
\end{tabular}
\caption{Point in time results for \system configurations with best-in-class $\locallm$ and $\remotelm$}
\label{tab:system-snapshot}
\end{table*}
We include additional experiment results from Section~\ref{subsec:results-model}. In Table~\ref{tab:remote-model-variations} we show the effects of varying $\remotelm$ on \system.  In Table~\ref{tab:system-snapshot}, we show the performance of \system using the best in-class models at the time (from late 2023 to late 2024).

\subsection{\naive $\locallm$ Analysis}
\begin{table}[h]
    \centering
    \begin{tabular}{cc}
        \toprule
        \textbf{Total Chunks In-Context} & \textbf{Accuracy} \\
        \midrule
        1   & 0.59375 \\
        16  & 0.53906 \\
        32  & 0.50000 \\
        64  & 0.48438 \\
        128 & 0.46094 \\
        \bottomrule
    \end{tabular}
    \caption{Accuracy vs. Number of Chunks in Context} Each chunk has 512 tokens.
    \label{tab:chunks_vs_accuracy}
\end{table}

\begin{table}[h]
    \centering
    \begin{tabular}{cc}
        \toprule
        \textbf{Number of Sub Tasks} & \textbf{Accuracy} \\
        \midrule
        1 & 0.70313 \\
        2 & 0.39844 \\
        3 & 0.19531 \\
        4 & 0.14844 \\
        \bottomrule
    \end{tabular}
    \caption{Accuracy vs. Number of Sub Tasks}
    \label{tab:subtasks_vs_accuracy}
\end{table}

\label{app:naive-analysis}

We perform an empirical analysis evaluating the robustness of $\locallm$. We perform experiments to evaluate two axes of model capabilities: (1) ability to reason over long contexts and (2) ability to solve multi-part queries. To test (1) and (2) we curate a synthetic dataset built over the \finance dataset wherein we use \gpt to construct an extraction based question-answering dataset over chunks (length 512 tokens) of documents in the \finance dataset. We then construct two settings evaluating over \llamathreetwo-3B-Instruct.
\\

\textbf{Long Context Reasoning}: To evaluate long-context reasoning, we concatenate between \{1,16,32,64,128\} chunks to construct the context. At least one chunk in the concatenated context contains the ground truth result. As seen in Table~\ref{tab:chunks_vs_accuracy}, increasing the context length from 512 to ~65.5K tokens leads to a 13 point drop in accuracy.
\\

\textbf{Multi-step Queries} To evaluate the ability of $\locallm$ to fulfill multi-step queries, we construct queries that have between \{1,2,3,4\} sub-tasks. Our results indicate increasing from 1 to sub-tasks leads to a 56.3 point drop in accuracy (see Table~\ref{tab:subtasks_vs_accuracy}).

\begin{table*}[]
\centering
\tiny

\begin{tabular}{p{0.7cm} p{0.7cm} p{0.7cm} p{0.4cm} p{0.4cm} p{0.4cm} p{0.4cm} p{0.4cm} p{0.4cm} p{0.4cm} p{0.4cm} p{0.4cm} p{0.4cm} p{0.4cm} p{0.4cm}}
\toprule
\multicolumn{1}{c}{\tiny \textbf{Protocol}} & \multicolumn{1}{c}{\tiny  \textbf{Local Model}} & \multicolumn{1}{c}{\tiny  \textbf{Remote Model}} & \multicolumn{4}{c}{\finance} & \multicolumn{4}{c}{\longhealth} & \multicolumn{4}{c}{\qasper} \\
 &  &  & Acc. & Cost & In Tok. (1k) & Out Tok. (1k) & Acc. & Cost & In Tok. (1k) & Out Tok. (1k) & Acc. & Cost & In Tok. (1k) & Out Tok. (1k) \\
\hline
Remote Only & --- & \gpt & 0.826 & \$0.261 & 103.04 & 0.32 & 0.748 & \$0.301 & 120.10 & 0.07 & 0.598 & \$0.137 & 54.40 & 0.09\\
\midrule
Local Only & \llamaeight & --- & 0.326 & \$0.000 & 0.00 & 0.00 & 0.468 & \$0.000 & 122.58 & 0.07 & 0.538 & \$0.000 & 54.41 & 0.06\\
Local Only & \llamaone & --- & 0.000 & \$0.000 & 0.00 & 0.00 & 0.115 & \$0.000 & 122.58 & 0.07 & 0.000 & \$0.000 & 54.41 & 0.10\\
Local Only & \llamathree & --- & 0.130 & \$0.000 & 0.00 & 0.00 & 0.345 & \$0.000 & 122.58 & 0.08 & 0.164 & \$0.000 & 54.41 & 0.08\\
Local Only & \qwenthree & --- & 0.087 & \$0.000 & 0.00 & 0.00 & 0.177 & \$0.000 & 31.24 & 0.08 & 0.156 & \$0.000 & 32.58 & 0.08\\
\midrule

\naive & \llamaeight & \gpt & 0.804 & \$0.007 & 0.88 & 0.46 & 0.635 & \$0.010 & 1.85 & 0.50 & 0.450 & \$0.007 & 0.92 & 0.42\\

\naive & \llamathree & \gpt & 0.698 & \$0.010 & 1.74 & 0.52 & 0.482 & \$0.009 & 1.56 & 0.47 & 0.372 & \$0.011 & 2.26 & 0.53\\

\naive & \qwenthree & \gpt & 0.217 & \$0.029 & 8.28 & 0.82 & 0.281 & \$0.021 & 5.70 & 0.68 & 0.210 & \$0.035 & 10.51 & 0.87\\
\midrule
\system & \llamaeight & \gpt & 0.804 & \$0.053 & 15.99 & 1.29 & 0.740 & \$0.054 & 18.96 & 0.65 & 0.582 & \$0.019 & 5.10 & 0.61\\
\system & \llamathree & \gpt & 0.726 & \$0.079 & 24.67 & 1.77 & 0.703 & \$0.057 & 20.11 & 0.66 & 0.558 & \$0.020 & 5.62 & 0.60\\
\system & \qwenthree & \gpt & 0.783 & \$0.059 & 17.20 & 1.56 & 0.645 & \$0.043 & 14.43 & 0.65 & -- & -- & -- & --\\
\system & \qwenseven & \gpt & -- & -- & -- & -- & -- & -- & -- & -- & 0.600 & \$0.015 & 3.44 & 0.61\\
\hline
\end{tabular}

\caption{\textbf{Accuracy and cost of local-remote systems.} Evaluation of cost and accuracy on the \finance \cite{islam2023financebench}, \longhealth\cite{adams2024longhealth}, and \qasper ~\cite{dasigi2021dataset}. The table compares two edge-remote communication protocols --- Naïve (\cref{sec:naive}) and \system (\cref{sec:methods}) --- alongside edge-only and remote-only baselines. 
Three different edge models are considered (\llamaeight, \llamathree, \qwenthree, and \llamaone) and a remote model (\gpt). 
Accuracy (Acc.) is the fraction of correct predictions across the dataset. 
Cost (USD) is the average cost in USD per query in the dataset computed. Costs are incurred for any calls to the remote model at \gpt rates (January 2025: \$2.50 per million input tokens and \$10.00 per million output tokens). We assume that running the edge model is free; see \cref{sec:prelim-setup} for details on the cost model.
In tokens is the number of input (\textit{i.e.} prefill) tokens sent to the remote model. 
Out tokens is the number of output (\textit{i.e.} decode) tokens generated from the remote model. Both values are shown in thousands. 
}
\label{table:app-tradeoff}
\end{table*}

\subsection{Relationship with Retrieval-Augmented Generation}\label{app:rag}
\begin{figure*}[t]
    \centering
    \includegraphics[width=1 \linewidth]{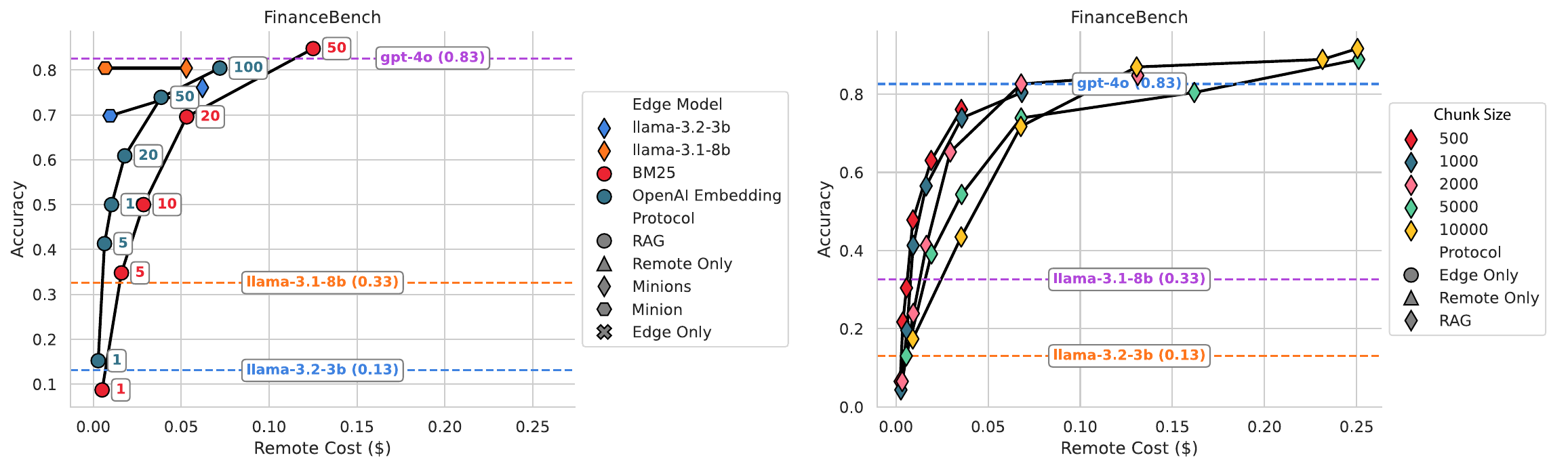}
    \vspace{-0.8cm}
    \caption{
    \textbf{Relationship with retrieval-augmented generation.} 
    }
    \label{fig:rag-comparison}
\end{figure*}

In this section, we discuss the relationship between local-remote collaboration and retrieval-augmented generation (RAG), a technique that reduces the number of tokens processed by an LM by retrieving a subset of relevant documents or chunks LM~\cite{lewis2020retrieval,karpukhin2020dense,lee2019latent}.

Retrieval-augmented generation and local-remote collaboration (\textit{e.g.} \system) are complementary techniques. 
They both provide a means to reduce cost by providing an LLM with a partial view of a large context. 
But, as we discuss below, they also have different error profiles and can be used in conjunction to improve performance.

\subsubsection{Comparison of \system and RAG on \finance}\label{app:rag:finance}
In \Cref{fig:rag-comparison} (left), we plot the quality-cost trade-off on \finance for local-remote systems (\naive and \system) and RAG systems using BM25 and OpenAI's \texttt{text-embedding-3-small} embeddings~\cite{article,neelakantan2022text}.
For RAG, we use a chunk size of 1000 characters, which we found to be optimal for this dataset after sweeping over chunk sizes with the BM25 retriever (see \Cref{fig:rag-comparison} (center)). We show how a simple hyperparameter (number of retrieved chunks provided to the remote model) allows us to trade off quality of the RAG system for remote cost.
Furthermore, we note that when the BM25 RAG system provides 50 or more chunks of the document to the remote model, it exceeds the performance of the remote model with the full context. This likely indicates that RAG helps in minimizing distractions from the long context. 
For \finance, when compared to \system, the RAG system with OpenAI embeddings reaches similar points in the quality-cost trade-off space. Interestingly however, none of the RAG configurations are are able to match the quality of \naive at the same low cost. 

\subsubsection{Comparison of \system and RAG (Embeddings + BM25) on Summarization Tasks}\label{app:rag:summary}
RAG is a very suitable approach for \finance, since all of the questions heavily rely on information extraction from specific sections of financial statements. However, RAG will not be suitable for a summarization task, unlike small LMs. Therefore, we use the long-document summarization dataset, \textsc{BooookScore}~\citep{chang2023booookscore}. \textsc{BooookScore} which contains a set of 400 books published between 2023-2024. The average story length in \textsc{Booookscore} is 128179 tokens with a max of 401486 tokens and a minimum of 26926 tokens. We utilize both \system, RAG (w/Embeddings + BM25), and \gpt only to complete the task. We describe the set-up for all three approaches next.
\\

\textbf{\system for summarization} In applying \system to the task, the $\locallm$ (\llamathreetwo-3B-Instruct) provides summaries on chunks of the original text, passing a list of chunk summaries to the $\remotelm$ (\gpt). $\remotelm$ produces the final summary. 

\textbf{RAG (Embedding) for summarization} In our embedding-based RAG approach, we use the OpenAI \textsc{text-embedding-3-small} to embed chunks of the original text (of length 5000 characters) and we retrieve the top-15 most relevant chunks using the query ``Summarize the provided text''. We then prompt \gpt to generate a complete summary over the retrieved chunks.

\textbf{RAG (BM25) for summarization} In our BM25-based RAG approach, we use the BM25 to retrieve chunks of the original text (of length 5000 characters) based on the query: ``Summarize the provided text''. We retrieve the top-15 most relevant chunks and prompt \gpt to produce a final summary over the retrieved chunks. We choose top-15 to ensure the number of tokens passed up by the baseline is comparable with those passed up by \system.

\textbf{GPT-4o} In our final baseline, we use \gpt alone to create the story summaries. For texts that extend beyond the 128K context length window, we truncate the stories.

\paragraph{Evaluation}
\begin{itemize}
    \setlength{\itemindent}{-1em}
    \item \textbf{Qualitative} In Table~\ref{tab:story-summaries} we provide samples outputs from each of the 4 methods described above. We highlight major events in red, themes in green, locations in blue and names in indigo. The samples demonstrate that amongst all the methods, \system outputs contain the most entity mentions and story specific details. Moreover, when compared to \gpt-only $\remotelm$,
\system is $9.3\times$ more efficient --- 11,500 versus the full 108,185 prefill tokens.

The summaries from \system are generally $1.3\times$ longer and more verbose than the RAG systems' summaries, likely indicating that the former is more effective at ``passing forward'' salient information. Moreover, RAG systems' summaries are missing the main arc of the narrative in favor of what seems an assortment of facts.  
    \setlength{\itemindent}{-1em}
    \item \textbf{Quantitative} We additionally perform a quantitative analysis of the generated summaries using a LLM-as-a-judge framework. As an evaluator, we use the \textsc{claude-3.5-sonnet} model, to avoid any biases between the evaluator and the supervisor model. We prompt the model with the generated summary, ground truth summary (gpt4-4096-inc-cleaned) provided from the original \textsc{BooookScore} generations, and a grading rubric (see Figure~\ref{rubric:summary}). The rubric evaluates 7 criteria: coherence, relevance, conciseness, comprehensiveness, engagement \& readability, accuracy, and thematic depth. We prompt \textsc{claude-3.5-sonnet} to generate a score (1-5) for each of the criteria and average the scores. We find that summaries generated by \system score comparably with \textsc{GPT4o}-only generated summaries, while RAG based baselines perform worse. Our results can be found in Table~\ref{tab:summary_qualitative_scores}.
\end{itemize}

\begin{figure}[h]
    \centering
    \begin{tcolorbox}[width=\textwidth, colframe=black!50, colback=white, sharp corners]
        \textbf{Evaluation Rubric for Summaries}
        \begin{enumerate}
            \item \textbf{Coherence (1-5):} Summary is logically structured, with clear connections between events, avoiding abrupt jumps or inconsistencies.
            \item \textbf{Relevance (1-5):} Accurately reflects key themes, events, and characters, focusing on essential details without unnecessary plot points.
            \item \textbf{Conciseness (1-5):} Thorough yet avoids excessive detail, presenting necessary information without redundancy.
            \item \textbf{Comprehensiveness (1-5):} Covers all major characters, events, and themes, ensuring a complete overview without omissions.
            \item \textbf{Engagement \& Readability (1-5):} Engaging and easy to read, with well-constructed sentences and clear, precise language.
            \item \textbf{Accuracy (1-5):} Stays true to the book’s storyline, themes, and tone, with correct details, names, and events.
            \item \textbf{Thematic Depth (1-5):} Identifies underlying themes and messages, providing insights into conflicts, motivations, and resolutions.
        \end{enumerate}
    \end{tcolorbox}
    \caption{Evaluation Rubric for Summaries}
    \label{rubric:summary}
\end{figure}

\begin{table}[h]
    \centering
    \begin{tabular}{lc}
        \hline
        Method & Score \\
        \hline
        \system & 3.01 \\
        GPT4o & 3.06 \\
        RAG (BM25) & 2.48 \\
        RAG (Embedding) & 2.38 \\
        \hline
        \hline
    \end{tabular}
    \caption{Comparison of Methods and Rubric Scores}
    \label{tab:summary_qualitative_scores}
\end{table}


\definecolor{character}{HTML}{4B0082}  
\definecolor{event}{HTML}{8B0000}      
\definecolor{theme}{HTML}{006400}      
\definecolor{location}{HTML}{000080}   

\scriptsize
\begin{longtable}{|p{1cm}|p{3cm}|p{3cm}|p{3cm}|p{3cm}|}    \caption{Comparative Analysis of Story Summaries Across Different Methods}
    \label{tab:story-summaries} \\
    \hline
    \textbf{Story Title} & \textbf{Minions} & \textbf{GPT-4o} & \textbf{BM25} & \textbf{Embedding} \\
    \hline
    \endfirsthead

    \hline
    \multicolumn{5}{|c|}{\textit{Continued from previous page}} \\
    \hline
    \textbf{Story Title} & \textbf{Minions} & \textbf{GPT-4o} & \textbf{BM25} & \textbf{Embedding} \\
    \hline
    \endhead

    \hline
    \multicolumn{5}{|r|}{\textit{Continued on next page}} \\
    \hline
    \endfoot

    \hline
    \endlastfoot

    \textit{A Likely Story} & The novel follows the complex dynamics of the {\color{character}Manning family}, focusing on {\color{character}Isabelle Manning}, a struggling writer grappling with the legacy of her famous father, {\color{character}Ward Manning}, and the recent death of her mother, {\color{character}Claire}. As Isabelle navigates her career and personal life, she uncovers family secrets, including her father's {\color{event}plagiarism} and her mother's {\color{event}hidden manuscript}, which challenge her identity and ambitions. The narrative explores themes of {\color{theme}fame, family expectations, and personal growth}, culminating in Isabelle's journey towards self-discovery and independence. Set against the backdrop of {\color{location}New York City} and {\color{location}Sag Harbor}, the story delves into the intricacies of familial relationships and the pursuit of creative fulfillment. & {\color{character}Isabelle Manning}, daughter of a famous author, {\color{event}discovers an unfinished manuscript} by her late mother, {\color{character}Claire}. She completes and publishes it as her own, leading to success but also guilt and exposure. Her father, {\color{character}Ward}, reveals he sabotaged her first book to teach her resilience, deepening her sense of betrayal. Ultimately, Isabelle confesses her actions, seeks redemption, and begins to redefine her identity beyond her father's shadow. & {\color{character}Isabelle Manning}, overshadowed by her famous father {\color{character}Ward}, receives a {\color{event}threatening text}, "I KNOW WHAT YOU DID," hinting at a secret tied to her book, "{\color{location}Underpainting}." Meanwhile, her mother {\color{character}Claire's} past support and her father's {\color{event}hidden illness} add layers to Isabelle's challenges. & In a narrative spanning {\color{location}New York} and {\color{location}Sag Harbor}, {\color{character}Claire Cunningham} grapples with her identity beyond being a {\color{location}Vassar} graduate, while her daughter {\color{character}Isabelle} faces personal and professional challenges, including her father's illness and her own writing struggles. \\
    \hline
    \textit{All the Dangerous Things} & {\color{character}Isabelle Drake}, a woman grappling with the {\color{event}traumatic disappearance of her son Mason}, navigates a complex web of grief, guilt, and suspicion. As she becomes entangled with true crime enthusiasts and investigators, including {\color{character}podcast host Waylon} and {\color{character}Detective Dozier}, Isabelle's quest for truth reveals unsettling family secrets and personal betrayals. Her journey is marked by strained relationships, particularly with her ex-husband {\color{character}Ben} and his connections to other women, including {\color{character}Valerie} and {\color{character}Allison}. Throughout the narrative, themes of {\color{theme}motherhood, mental health, and societal judgment} are explored, culminating in a deeper understanding of the pressures and expectations faced by women. & {\color{character}Isabelle Drake}, plagued by {\color{theme}insomnia and guilt}, is desperate to find her missing son, {\color{character}Mason}. She suspects her husband, {\color{character}Ben}, and his new partner, {\color{character}Valerie}. With {\color{character}Waylon's} help, she discovers {\color{character}Abigail Fisher}, manipulated by {\color{character}Valerie}, took Mason believing she was rescuing him. & The narrative follows {\color{character}Isabelle}, dealing with {\color{event}Mason's disappearance}. She works with podcaster {\color{character}Waylon}, uncovering links to {\color{character}Ben's} deceased wife, {\color{character}Allison}. & {\color{character}Isabelle}, struggling with {\color{theme}grief and insomnia}, joins a {\color{location}grief counseling group}. She meets {\color{character}Valerie} and collaborates with {\color{character}Waylon}, but becomes wary after finding unsettling information on his laptop. \\
    \hline
    \textit{A Living Remedy: A Memoir} & {\color{character}Nicole Chung}, a {\color{theme}Korean American adoptee}, reflects on her complex relationships with her adoptive parents, her identity, and the challenges of navigating life as a minority in a predominantly white community in {\color{location}Oregon}. Her memoir explores themes of {\color{theme}family, loss, and resilience}, as she recounts her {\color{event}father's death from kidney failure}, and her {\color{event}mother's battle with cancer}. Amidst these personal challenges, Chung grapples with her own grief, financial struggles, and the impact of the {\color{event}COVID-19 pandemic}, while finding solace in her family, faith, and writing. Her journey is marked by a deep appreciation for her parents' sacrifices, the support of her husband and children, and the enduring legacy of love and forgiveness instilled by her mother. & {\color{character}Nicole Chung's} memoir explores her journey after the {\color{event}loss of her adoptive parents}. As a Korean adoptee, she reflects on family's financial struggles, parents' health battles, and their deaths' impact on her identity. She finds solace in writing and her own family. & The protagonist struggles with {\color{event}visiting her dying mother during the COVID-19 pandemic}. The story explores {\color{theme}grief, family responsibility, and cherishing life} amidst adversity. & A woman reflects on her {\color{event}parents' illnesses and deaths}, balancing her role as a daughter and mother. She finds solace in {\color{theme}childhood memories} and the legacy of her parents' love. \\
    \hline
        \textit{A House with Good Bones} & {\color{character}Samantha}, a 32-year-old archaeoentomologist, returns to her childhood home on {\color{location}Lammergeier Lane} in {\color{location}North Carolina}, where she confronts her family's dark past, including her grandmother {\color{character}Gran Mae's} mysterious and malevolent legacy. As Samantha navigates her mother's strange behavior and the {\color{event}eerie presence of vultures}, she uncovers secrets involving {\color{event}ritual magic}, a jar of human teeth, and the supernatural "underground children." With the help of her friend {\color{character}Gail} and handyman {\color{character}Phil}, Samantha faces the haunting manifestations of her family's history. The novel explores themes of {\color{theme}family, memory, and the supernatural}, blending elements of horror and fantasy. & {\color{character}Samantha Montgomery} returns home to find her mother acting strangely and the house devoid of insects. She uncovers a dark history involving her {\color{character}great-grandfather}, a sorcerer, and her grandmother, who used {\color{event}roses to wield power}. With help from {\color{character}Gail} and {\color{character}Phil}, she confronts the terrifying "{\color{event}underground children}," using rose power to banish threats. & The protagonist returns to their {\color{character}grandmother's} unchanged garden, filled with roses but {\color{event}mysteriously devoid of insects}. They uncover {\color{theme}unsettling truths about their grandmother's past} and their mother's current state of mind. The narrative explores themes of {\color{theme}family legacy and the passage of time}. & {\color{character}Samantha}, an archaeoentomologist, returns to her childhood home and finds herself investigating {\color{event}insect collections}. Dealing with {\color{theme}sleep paralysis} and memories of her grandmother, she discovers the {\color{event}peculiar absence of insects} in the garden. She navigates family dynamics and her mother's anxiety amid an eerie atmosphere. \\

\end{longtable}

\section{Prompts}
\label{app:methods-prompts}

\subsection{\naive}

\textbf{$\remotelm$}

\begin{tcolorbox}[colback=gray!10,  width=\textwidth]
\begin{lstlisting}[breaklines]
We need to answer the following question based on a {doc_type}.

### Question
{query}

### Instructions
You will not have direct access to the {doc_type}, but can chat with a small language model which has read the entire thing.

Feel free to think step-by-step, but eventually you must provide an output 
in the format below:

<think step by step here>
```json
{{
    "message": "<your message to the small language model>"
}}
```
\end{lstlisting}
\end{tcolorbox}

\textbf{$\locallm$}

\begin{tcolorbox}[colback=gray!10,  width=\textwidth]
\begin{lstlisting}[breaklines]
You will help a user answer the following question based on a {doc_type}. 


Read the {doc_type} below and prepare to answer questions from an expert user. 
### {doc_type}
{context}

### Question
{query}
\end{lstlisting}
\end{tcolorbox}

\textbf{$\mathrm{Conversation}$}

\begin{tcolorbox}[colback=gray!10,  width=\textwidth]
\begin{lstlisting}[breaklines]
Here is the response from the small language model:

### Response
{response}


### Instructions
Analyze the response and think-step-by-step to determine if you have enough 
information to answer the question.

If you have enough information, provide a final numeric answer in the format
below.

<think step by step here>
```json
{{
    "decision": "provide_final_answer",  
    "answer": "<your answer>"
}}
```

Otherwise, request additional information from the small language model by 
outputting the following:

<think step by step here>
```json
{{
    "decision": "request_additional_info",
    "message": "<your message to the small language model>"
}}
```
\end{lstlisting}
\end{tcolorbox}

\subsection{\system}


\paragraph{\system: \finance} \

\textbf{$\mathrm{Decompose}$}

\begin{tcolorbox}[colback=gray!10,  width=\textwidth]
\begin{lstlisting}[breaklines]
# Decomposition Round #{step_number}

You do not have access to the raw document(s), but instead can assign tasks to small and less capable language models that can read the document(s).
Note that the document(s) can be very long, so each task should be performed only over a small chunk of text.

Write a Python function that will output formatted tasks for a small language model.
Make sure that NONE of the tasks require calculations or complicated reasoning.
Any information you mentioned in your task should be given an extraction task.

Please use chunks of {pages_per_chunk} pages using the `chunk_on_multiple_pages(doc = context, pages_per_chunk ={pages_per_chunk})` function.

If you have multiple tasks, consider using nested for-loops to apply a set of tasks to a set of chunks. Though it's not required to have more than one task.

{ADVANCED_STEPS_INSTRUCTIONS}

Assume a Pydantic model called `JobManifest(BaseModel)` is already in global scope. For your reference, here is the model:
```
{manifest_source}
```
Assume a Pydantic model called `JobOutput(BaseModel)` is already in global scope. For your reference, here is the model:
```
{output_source}
```
DO NOT rewrite or import the model in your code.

The function signature will look like:
```
{signature_source}
```

You can assume you have access to the following chunking function(s). Do not reimplement the function, just use it.
```
{chunking_source}
```
\end{lstlisting}
\end{tcolorbox}

\textbf{$\mathrm{Worker}$}

\begin{tcolorbox}[colback=gray!10,  width=\textwidth]
\begin{lstlisting}[breaklines]
Your job is to complete the following task using only the context below. The context is a chunk of text taken arbitrarily from a document, it might or might not contain relevant information to the task.

## Document
{context}

## Task
{task}

{advice}

Return your result in JSON with the following keys: "explanation", "citation", and "answer".

- "explanation": A concise statement of your reasoning or how you concluded your answer.
- "citation": A direct snippet of the text that supports your answer. If nothing is found, put "None".
- "answer": The extracted answer. If nothing is found, put "None".

Be certain to only rely on the provided text. If you cannot determine the information confidently from this chunk, respond with "None" for all fields.
\end{lstlisting}
\end{tcolorbox}

\textbf{$\mathrm{Synthesize}$}
\begin{tcolorbox}[colback=gray!10,  width=\textwidth]
\begin{lstlisting}[breaklines]

Now synthesize the findings from multiple junior workers (LLMs). 
Your task is to finalize an answer to the question below **if and only if** you have sufficient, reliable information. 
Otherwise, you must request additional work.

---
## Inputs
1. Question to answer:
{question}

2. Collected Job Outputs (from junior models):
{extractions}

---
First think step-by-step and then answer the question using the exact format below.

## ANSWER GUIDELINES
1. **Determine if the collected Job Outputs provide enough trustworthy, consistent evidence to confidently answer the question.** 
   - If the data is incomplete or contradictory, do NOT guess. Instead, specify what is missing.
   - If the evidence is sufficient, provide a final answer.

2. **Be conservative.** When in doubt, ask for more information.

3. **Address conflicts.** If multiple jobs give different answers, rely on whichever is best supported by a valid "explanation" and "citation".
   - If you need more information from the conflicting jobs you could request additional work from those specific jobs (be sure to mention the specific job IDs in your additional_info field).
   - Then, in the next round you can make a smaller set of jobs to determine which answer is correct.

4. **Required JSON Output**: You must output a JSON object with these keys:
   - "decision": Must be either "provide_final_answer" OR "request_additional_info".
     - Use "provide_final_answer" if you have enough information.
     - Use "request_additional_info" if you cannot conclusively answer.
   - "explanation": A short statement about how you arrived at your conclusion or what is still missing.
   - "answer": The final answer string if "decision"="provide_final_answer", or null otherwise. Should contain ONLY the final answer, without additional calculations or explanations.
     
Here is the template for your JSON response (with no extra text outside the JSON):

<think step-by-step here>
```json
{{
"decision": "...",
"explanation": "...",
"answer": "... or null", # Good answer format: "0.56"; Bad answer format: "The ratio is calculated as 1-0.27*2 = 0.56"
}}
```

**Important**:
- If there is not enough information, set "answer" to null, set "decision" to "request_additional_info", and specify exactly what else you need in "missing_info" and from which job IDs.

Now, carefully inspect the question, think step-by-step and perform any calculations before outputting the JSON object.
\end{lstlisting}
\end{tcolorbox}

\paragraph{\system: \longhealth} \

\textbf{$\mathrm{Decompose}$}
\begin{tcolorbox}[colback=gray!10,  width=\textwidth]
\tiny
\begin{lstlisting}[breaklines]
# Decomposition Round #{step_number}

You do not have access to the raw document(s), but instead can assign tasks to small and less capable language models that can read the document(s).
Note that the document(s) can be very long, so each task should be performed only over a small chunk of text. 

Write a Python function that will output formatted tasks for a small language model.
Make sure that NONE of the tasks require multiple steps. Each task should be atomic! 
Consider using nested for-loops to apply a set of tasks to a set of chunks.
The same `task_id` should be applied to multiple chunks. DO NOT instantiate a new `task_id` for each combination of task and chunk.
Use the conversational history to inform what chunking strategy has already been applied.

{ADVANCED_STEPS_INSTRUCTIONS}

Assume a Pydantic model called `JobManifest(BaseModel)` is already in global scope. For your reference, here is the model:
```
{manifest_source}
```
Assume a Pydantic model called `JobOutput(BaseModel)` is already in global scope. For your reference, here is the model:
```
{output_source}
```
DO NOT rewrite or import the model in your code.

The function signature will look like:
```
{signature_source}
```


You can assume you have access to the following chunking function(S). Do not reimplement the function, just use it.
```
{chunking_source}
```

Here is an example
```
task_id = 1  # Unique identifier for the task
for doc_id, document in enumerate(context):
    # if you need to chunk the document into sections
    chunks = chunk_by_section(document)
    # or if you need to chunk the document into pages
    chunks = chunk_by_page(document)

    for chunk_id, chunk in enumerate(chunks):
        # Create a task for extracting mentions of specific keywords
        task = (
            "Extract all mentions of the following keywords: "
            "'Ca19-9', 'tumor marker', 'September 2021', 'U/ml', 'Mrs. Anderson'."
        )
        job_manifest = JobManifest(
            chunk_id=f"doc_id_chunk_id",
            task_id=task_id,
            chunk=chunk,
            task=task,
            advice="Focus on extracting the specific keywords related to Mrs. Anderson's tumor marker levels."
        )
        job_manifests.append(job_manifest)
```
\end{lstlisting}
\end{tcolorbox}

\textbf{$\bp_{\text{worker}}$}

\begin{tcolorbox}[colback=gray!10,  width=\textwidth]
\begin{lstlisting}[breaklines]
Your job is to complete the following task using only the context below. The context is a chunk of text taken arbitrarily from a document, it might or might not contain relevant information to the task.

## Document
{context}

### Question you are trying to answer: 
{question}

# You have been instructed to extract information pertaining to the following concepts: 
# \"Date of visit\", {task}

Format your response as follows:
{{
"Date of visit" : "`direct quote extracted text`",
"<keyword_1>" : "`direct quote extracted text`", 
"<keyword_2>" : "`direct quote extracted text`", 
...
}}

Can you please extract the relevant sections from the document that are related to the concepts provided? Extract direct quotes or sentences. If concept is not mentioned, leave it out.

Your Answer:
\end{lstlisting}
\end{tcolorbox}

\textbf{$\bp_{\text{synthesize}}$}

\begin{tcolorbox}[colback=gray!10,  width=\textwidth]
\begin{lstlisting}[breaklines]
Answer the following by the synthesizing findings from multiple junior workers (LLMs).


---
## Inputs
1. Question to answer:
{question}

2. Collected Job Outputs (from junior models):
{extractions}

---
First think step-by-step and then answer the question using the exact format below.

## ANSWER GUIDELINES

**Required JSON Output**: You must output exactly one JSON object with these keys:
   - "decision": Must be  "provide_final_answer".
   - "explanation": A short statement about how you arrived at your conclusion or what is still missing.
   - "answer": The final answer string (that matches one of the provided options) if "decision"="provide_final_answer", or null otherwise.

     
Here is the template for your JSON response:

<think step-by-step here>


{{
"decision": "...",
"explanation": "...",
"answer": "...",
}}


Now, carefully inspect the question, think step-by-step and perform any calculations before outputting the JSON object. If answer choices are provided, your answer must **exactly** match one of the answer choices.

Question: 
{question}

Your Answer:
\end{lstlisting}
\end{tcolorbox}

\paragraph{\system: \qasper} \

\textbf{$\bp_{\text{decompose}}$}
\begin{tcolorbox}[colback=gray!10,  width=\textwidth]
\begin{lstlisting}[breaklines]
# Decomposition Round #{step_number}

You do not have access to the raw document(s), but instead can assign tasks to small and less capable language models that can read the document(s).
Note that the document(s) can be very long, so each task should be performed only over a small chunk of text. 

Write a Python function that will output formatted tasks for a small language model.
Make sure that NONE of the tasks require multiple steps. Each task should be atomic! 
Consider using nested for-loops to apply a set of tasks to a set of chunks.
The same `task_id` should be applied to multiple chunks. DO NOT instantiate a new `task_id` for each combination of task and chunk.
Use the conversational history to inform what chunking strategy has already been applied.

{ADVANCED_STEPS_INSTRUCTIONS}

Assume a Pydantic model called `JobManifest(BaseModel)` is already in global scope. For your reference, here is the model:
```
{manifest_source}
```
Assume a Pydantic model called `JobOutput(BaseModel)` is already in global scope. For your reference, here is the model:
```
{output_source}
```
DO NOT rewrite or import the model in your code.

The function signature will look like:
```
{signature_source}
```


You can assume you have access to the following chunking function(S). Do not reimplement the function, just use it.
```
{chunking_source}
```

Here is an example
```
task_id = 1  # Unique identifier for the task
for doc_id, document in enumerate(context):
    # if you need to chunk the document into sections
    chunks = chunk_by_section(document)
    # or if you need to chunk the document into pages
    chunks = chunk_by_page(document)

    for chunk_id, chunk in enumerate(chunks):
        # Create a task for extracting mentions of specific keywords
        task = (
            "Extract all mentions of the following keywords: "
            "'Ca19-9', 'tumor marker', 'September 2021', 'U/ml', 'Mrs. Anderson'."
        )
        job_manifest = JobManifest(
            chunk_id=f"doc_id_chunk_id",
            task_id=task_id,
            chunk=chunk,
            task=task,
            advice="Focus on extracting the specific keywords related to Mrs. Anderson's tumor marker levels."
        )
        job_manifests.append(job_manifest)
```
\end{lstlisting}
\end{tcolorbox}

\textbf{$\bp_{\text{worker}}$}

\begin{tcolorbox}[colback=gray!10,  width=\textwidth]
\begin{lstlisting}[breaklines]
Your job is to complete the following task using only the context below. The context is a chunk of text taken arbitrarily from a document, it might or might not contain relevant information to the task.

## Document
{context}

### Question you are trying to answer: 
{question}

# You have been instructed to extract information pertaining to the following concepts: 
# \"Date of visit\", {task}

Format your response as follows:
{{
"Date of visit" : "`direct quote extracted text`",
"<keyword_1>" : "`direct quote extracted text`", 
"<keyword_2>" : "`direct quote extracted text`", 
...
}}

Can you please extract the relevant sections from the document that are related to the concepts provided? Extract direct quotes or sentences. If concept is not mentioned, leave it out.

Your Answer:
\end{lstlisting}
\end{tcolorbox}

\textbf{$\bp_{\text{synthesize}}$}

\begin{tcolorbox}[colback=gray!10,  width=\textwidth]
\begin{lstlisting}[breaklines]
Answer the following by the synthesizing findings from multiple junior workers (LLMs).


---
## Inputs
1. Question to answer:
{question}

2. Collected Job Outputs (from junior models):
{extractions}

---
First think step-by-step and then answer the question using the exact format below.

## ANSWER GUIDELINES

**Required JSON Output**: You must output exactly one JSON object with these keys:
   - "decision": Must be "provide_final_answer" or "need more information"
   - "explanation": A short statement about how you arrived at your conclusion or what is still missing.
   - "answer": a final answer that is a text span pulled directly from the job output citations.

     
Here is the template for your JSON response:

<think step-by-step here>

{{
"decision": "...",
"explanation": "...",
"answer": "..,", 
}}

Now, carefully inspect the question, think step-by-step and perform any calculations before outputting the JSON object. 
- If answer choices are provided, your answer must **exactly** match one of the answer choices.
- Don't paraphrase the final answer --- extract text directly from the document(s) or previous job outputs.

Question: 
{question}

Your Answer:
\end{lstlisting}
\end{tcolorbox}





\end{document}